\documentclass[manuscript,screen]{acmart}

\AtBeginDocument{%
  \providecommand\BibTeX{{%
    \normalfont B\kern-0.5em{\scshape i\kern-0.25em b}\kern-0.8em\TeX}}}

\usepackage{amsmath}

\usepackage{multirow}

\newcommand\toolss{27}

\begin{document}

\title{Location reference recognition from texts: A survey and comparison}


\author{Xuke Hu}
\authornote{Corresponding author}
\email{xuke.hu@dlr.de}
\orcid{1234-5678-9012}

\affiliation{%
  \institution{Institute of Data Science, German Aerospace Center (DLR)}
  %\streetaddress{}
  \city{Jena}
 % \state{}
  \country{Germany}
 % \postcode{}
}

\author{Zhiyong Zhou}
\authornote{Corresponding author}
\email{zhiyong.zhou@geo.uzh.ch}
\orcid{1234-5678-9012}

\affiliation{%
  \institution{Department of Geography, University of Zurich}
%  \streetaddress{}
  \city{Zurich}
 % \state{}
  \country{Switzerland}
 % \postcode{}
}

\author{Hao Li}
\email{hao.li@uni-heidelberg.de}
\orcid{1234-5678-9012}

\affiliation{%
  \institution{Department of Geography, University of Heidelberg}
%  \streetaddress{}
  \city{Heidelberg}
 % \state{}
  \country{Germany}
 % \postcode{}
}

\author{Yingjie Hu}
\email{yhu42@buffalo.edu}
\orcid{1234-5678-9012}

\affiliation{%
  \institution{Department of Geography, University at Buffalo}
%  \streetaddress{}
  \city{Buffalo}
 % \state{}
  \country{USA}
 % \postcode{}
}

\author{Fuqiang Gu}
\email{gufq@cqu.edu.cn}
\orcid{1234-5678-9012}

\affiliation{%
  \institution{College of Computer Science, Chongqing University}
%  \streetaddress{}
  \city{Chongqing}
%  \state{}
  \country{China}
 %% \postcode{}
}

\author{Jens Kersten}
\email{Jens.Kersten@dlr.de}
\orcid{1234-5678-9012}

\affiliation{%
  \institution{Institute of Data Science, German Aerospace Center (DLR)}
 % \streetaddress{}
  \city{Jena}
%  \state{}
  \country{Germany}
 % \postcode{}
}

\author{Hongchao Fan}
\email{hongchao.fan@ntnu.no}
\orcid{1234-5678-9012}

\affiliation{%
  \institution{Department of Civil and Environmental Engineering, Norwegian
University of Science and Technology}
%  \streetaddress{}
  \city{Trondheim}
%  \state{}
  \country{Norway}
 % \postcode{}
}

\author{Friederike Klan}
\email{Friederike.Klan@dlr.de}
\orcid{1234-5678-9012}

\affiliation{%
  \institution{Institute of Data Science, German Aerospace Center (DLR)}
 % \streetaddress{}
  \city{Jena}
%  \state{}
  \country{Germany}
 % \postcode{}
}

\renewcommand{\shortauthors}{Hu, et al.}

\begin{abstract}

	A vast amount of location information exists in unstructured texts, such as social media posts, news stories, scientific articles, web pages, travel blogs, and historical archives. Geoparsing refers to the process of recognizing location references from texts and identifying their geospatial representations. While geoparsing can benefit many domains, a summary of the specific applications is still missing. Further, there lacks a comprehensive review and comparison of existing approaches for location reference recognition, which is the first and a core step of geoparsing. To fill these research gaps, this review first summarizes seven typical application domains of geoparsing: geographic information retrieval, disaster management, disease surveillance, traffic management, spatial humanities, tourism management, and crime management. We then review existing approaches for location reference recognition by categorizing these approaches into four groups based on their underlying functional principle: rule-based, gazetteer matching-based, statistical learning-based, and hybrid approaches. Next, we thoroughly evaluate the correctness and computational efficiency of the \toolss\thinspace most widely used approaches for location reference recognition based on 26 public datasets with different types of texts (e.g., social media posts and news stories) containing 39,736 location references across the world. Results from this thorough evaluation can help inform future methodological developments for location reference recognition, and can help guide the selection of proper approaches based on application needs. %Last, we analyzed the challenge and future research directions.
%, including their performance on formal and informal texts and on places of different attributes (e.g., category, length, and form) and their computational efficiency
\end{abstract}



 \maketitle
	\section{Introduction}
	\label{Section: Introduction}
	
\begin{figure}[htbp!]
		\centering
		\includegraphics[width=0.8\textwidth]{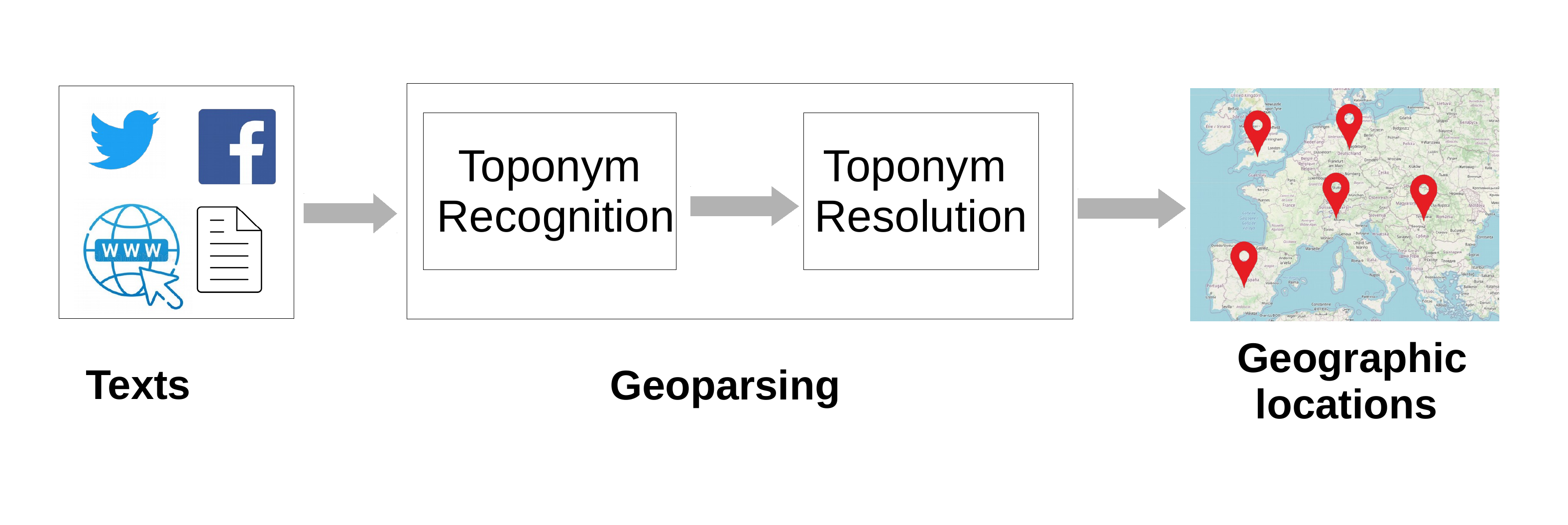}
		\caption{The general workflow of geoparsing and its two steps.} %The dashed arrow represents that the step  is optional.}
		\label{workflow_figure}
	\end{figure}

	"Location matters, and not just for real estate" \citep{wallgrun2018geocorpora}. %In context-aware computing, location is a fundamental component that supports a wide range of applications \citep{hazas2004location}. 
	% It was stated by \citet{huxhold1991introduction} that `\textit{Eighty to ninety percent of all the information collected and used was related to geography}'. %A similar statement appeared in \citep{williams1987selling} that \textit{`It has been estimated that approximately 80\% of the informational needs of a local government policymaker is related to a geographical location'}.
	With the rapid development of the Global Navigation Satellite System (GNSS), sensor-rich (e.g., inertial sensors, Wi-Fi module, and cameras) smart devices, and ubiquitous communication infrastructure (e.g., cellular and 4G networks and Wi-Fi access points), our capability of obtaining location information of moving objects and events in both indoor and outdoor spaces has been dramatically improved \citep{shang2015improvement}. This increased our  ability to better understand geo-spatial processes and to support decision making in all contexts from business, entertainment, to crisis management \citep{wallgrun2018geocorpora}. Apart from sensor equipment, natural language texts (e.g., social media posts, web pages, and news stories) are another important source that contains much geospatial information in the form of \textit{location references}. These \textit{location references} embedded in texts can be in the form of simple \textit{place names} (or \textit{toponyms}), and can also be in the form of \textit{location descriptions} that contain both place names and additional spatial modifiers (e.g., direction,  distance, and spatial relationship) \citep{vasardani2013locating}.   %For simplicity, throughout this work we use the terms \textit{place name}, \textit{location reference}, and \textit{toponym} interchangeable. 
	\textit{Geoparsing} refers to the process of recognizing location references from texts and identifying their geospatial representations. \textit{Geoparsing} is an ongoing research problem that has been studied over the past two decades \citep{jones2002spatial,amitay2004web,silva2006adding,aldana2020adaptive,hu2020GazPNE}. %Geoparsing can be divided into three types based on the level of resolution: (1) toponym-level \citep{middleton2018location,al2017location, amitay2004web,karimzadeh2019geotxt} that infers geographical location of toponyms, (2) event-level \citep{halterman2018linking,dewandaru2020event,imani2017focus} that infers geographical location of events mentioned in texts, and (3) document-level \citep{woodruff1994gipsy, lieberman2007steward, silva2006adding, andogah2012every,purves2007design} that infers single geographic focus of documents. The geoparser at the toponym-level 
	It consists of two steps: (1) toponym recognition, which is also called location reference recognition (2) toponym resolution, which is also called geocoding that disambiguates toponyms and identifies their geographic coordinates. %Event-level and document-level geoparsers are built on toponym-level geoparsers, including the above two steps. Event-level geoparser needs to further link each toponym to an event mentioned in the text or a null event, while document-level geoparser needs to determine the geographic scope of a document based on resolved toponyms. 
	%The task of geoparsing is to extract toponyms from texts and encode them in coordinates \citep{aldana2020adaptive}, while the task of GSE is to predict the location to which an entire text/ document is geographically relevant. In many approaches for GSE, geoparsing is the subtask of GSE \citep{woodruff1994gipsy, alexopoulos2012optimizing,alexopoulosklocator}. That is, GSE normally consists of three steps: toponym recognition, toponym resolution, and grounding toponyms that performs a polygonal overly of the data output in the second step. 
	Figure \ref{workflow_figure} illustrates the workflow of geoparsing.
		 % A system capable of geoparsing can enable various applications, such as geographical information retrieval \citep{freire2011metadata,purves2018geographic},  %knowledge discovery \citep{tamames2010envmine, weissenbacher2015knowledge, acheson2021extracting},
	% disaster management \citep{lingad2013location,shook2016socio}, %emergency response \citep{cresci2017nowcasting,ozdikis2017survey,de2018taggs,singh2019event}, early warning \citep{avvenuti2016predictability}, 
%	 traffic management \citep{he2013improving,liu2016collective,milusheva2021applying}, touristic planning \citep{colladon2019using,cresci2014towards,brilhante2015planning}, %healthcare accessibility \citep{li2017designing}, 
%	 disease surveillance \citep{lampos2012nowcasting, gritta2019you, yang2012global},  crime prevention \citep{arulanandam2014extracting,vomfell2018improving,dasgupta2017crimeprofiler,das2019graph,po2018building}, and spatial humanity \citep{rupp2013customising, gregory2015geoparsing, won2018ensemble, tateosian2017tracking}. These applications will be explained in Section \ref{application}.

%or more by .
	
%	which was initially and extensively investigated in the field of geographic information retrieval (GIR) .
	
 %The textual domains, in which the geoparsing approaches have been implemented, can be divided into two types: formal and informal texts. 
 
	% The amount of the digital textual material available to researchers is proliferating rapidly. Some of the major digital collections of historical books and reports include the , which each comprise many millions or even billions of words. 
	
% , and retrieve articles using spatial search criteria Last but not least, from published scientific articles, fine-grained geographical information about where a certain virus was first reported and how did it spread geographically, can be automatically extracted, which were normally missing in GenBank records. %By extracting the geographic locations and damage caused by natural disasters from hundreds of news articles, the cost of the disaster at certain regions can be automatically summarized. 
 %By performing recognition and resolution of place names mentioned over the descriptive metadata records of typical digital libraries, user experience in resource discovery is expected to improve since the geocoded geographic entities can help users to evaluate if a particular resource is relevant for his information need \citep{freire2011metadata}.

	Geoparsing has traditionally been used in formal texts for location extraction, such as web pages, news, scientific articles, travel blogs, and historical archives \citep{amitay2004web,wallgrun2018geocorpora}. However, the drastically increased importance of social media data (SMD) in various domains such as social science, policy making, and humanitarian relief \citep{tapia2013beyond, harris2014communication, avvenuti2018crismap,curiel2020crime} has facilitated efforts to extend geoparsing to informal texts \citep{wallgrun2018geocorpora}. According to Statista \footnote{\url{https://www.statista.com/statistics/278414/number-of-worldwide-social-network-users/}}, the number of worldwide social network users will reach 4.4 billion by 2025. On average, 500 million tweets \footnote{\url{https://www.dsayce.com/social-media/tweets-day/}} and 4.75 billion Facebook posts \footnote{\url{https://blog.wishpond.com/post/115675435109/40-up-to-date-facebook-facts-and-stats}} are sent each day. %Social media data has thus become one of the most useful sources for gathering situational information in real-time due to their extensive use, speed, and coverage  \citep{tapia2013beyond,reuter2018fifteen}. %For example, to aid emergency response of disasters, decision support systems \citep{lingad2013location,shook2016socio} have been proposed, which demand the location-enabled tweets related to disasters to timely map and visualize the situational information, such as the disaster phrase (e.g., preparedness, impact, response, and recovery) \citep{wang2020tracking}, rescue request \citep{singh2019event, zou2021social}, resource (e.g., food, clothing, water, medical treatment, and shelter) needs and availabilities \citep{basu2022utilizing, dutt2019utilizing}, and facility status (e.g., damaged, closed, or reopened) \citep{fan2020hybrid,belcastro2021using,belcastro2021using} in the aftermath of mass disasters. With the crisis map, the first responders can track the unfolding situation and identify stricken locations that require prioritized intervention \citep{avvenuti2018gsp} and realize optimized real-time resource allocation \citep{singh2019event}, the government can conduct the damage assessment of the disasters \citep{wu2018disaster} in a faster manner, and the public can search for the locations where they can obtain a certain resource. %Twitter and Weibo are widely used by both public and government in disaster response. 
% Explicit geographic information is not only needed in emergency responses \citep{cresci2017nowcasting,ozdikis2017survey,de2018taggs,singh2019event}, but also in epidemics monitoring \citep{lampos2012nowcasting,dredze2013carmen,zhu2021using}. The epidemic of coronavirus disease 2019 (COVID-19) has become a severe public health crisis \citep{sohrabi2020world}. After its happening, geotagged tweets were utilized to analyze the mental health status of the public after the occurrence of COVID-19 \citep{zhu2021using,hohl2021understanding}, to track and visualize the spread and diffusion of COVID-19 \citep{andreadis2021social}, and to reveal human mobility dynamics during the COVID-19 pandemic \citep{iranmanesh2021reading, huang2020twitter}.  %For instance, when an emergency event occurs, such as earthquakes, floods, fire, terrorist attacks, and civil unrest \citep{alexander2014social,ozdikis2017survey,yuan2018feasibility}, the location information in microblogs is valuable for keeping people and authorities informed about the exact affected area, where people can receive fresh water and food, and the locations where people need rescue and medical assistance \citep{kar2018d, maneriker2019pipeline}. 
Formal texts normally do not have location-related metadata, while informal texts, such as tweets, can be geotagged, i.e., a Twitter user can select a location and attach that location to the posted message. However, geotagged tweets are rare, %According to Message Understanding Conference (MUC) campaigns \footnote{\url{http://wwwJtl.nist.gov/iaui/894.02/related projects/tipster/muc.htm/}}, events have several dimensions that are equally important and require specific attention. The main dimensions are three folds: location information that indicates where an event has happened, temporal information that indicates when an event has happened, and entity-related information that indicates what the event is about \citep{hoang2018location}. This study focuses on the location dimension.
and according to \citet{cheng2010you}, \citet{morstatter2013sample}, and \citet{kumar2017authenticity}, only 0.42\%, 3.17\%, and 7.90\% of the total number of tweets contain geotags, respectively. %Important situational information published by government offices, organizations, news media, and reporters normally do not include geotags, while individuals may not attach geotags to their tweets due to privacy concerns. 
In addition, Twitter removed their precise geotagging feature in June 2019, showing only a rough location, e.g., the bounding box of a tagged place rather than a pair of latitude and longitude coordinates. This change could lead to a further decrease in the number of geotagged tweets    \citep{hu2020understanding}. %For instance, the tweet “Will discuss on Times Now at 8.30 am today regarding Dengue Fever in Tamil Nadu.” clearly refers to Tamil Nadu, but the geo-tagged location is New Delhi (from where the tweet was posted).
  The geotagged locations of tweets are not always the same as the locations described in their tweet content either \citep{lingad2013location}. %Conversely, according to \citep{gazpne2}, over half of informative tweets during emergencies contain at least one location reference, which refer to admin units, such as states and cities, and finer and more precise locations, such as streets, intersections, churches, schools, and home addresses. %Therefore, many studies used the mentioned location in tweets to  infer the tweet location where a tweet is posted.
  %Moreover, according to \citet{vieweg2010microblogging}, location references in tweets is usually the location needed for monitoring and or response to an emergency. 
  In a nutshell, it is often necessary to extract location references from unstructured texts. Notably, informal texts, such as tweets, are short, have few or no formatting or grammatical requirements, and can have uncommon abbreviations, slang, and misspellings, which pose additional challenges for geoparsing \citep{wang2019we}.
While there exist quite some studies on geoparsing \citep{gritta2018s,purves2018geographic}, we identify two gaps in the literature that motivate this current review paper. First, the many possible applications of geoparsing are scattered in individual papers \citep{amitay2004web,abdelkader2015brands,gelernter2013algorithm,gregory2015geoparsing} or are only partially reviewed \citep{hu2021harvesting, gritta2020pragmatic}, and there lacks a systematic and more comprehensive summary of these applications. Consequently, it is difficult for researchers who are new to geoparsing to have a quick view of these many possible applications. Second, existing review papers on geoparsing, such as \citep{melo2017automated, monteiro2016survey,gritta2018s,wang2019enhancing}, focused on the entire workflow of geoparsing (i.e., both of the two steps) rather than location reference recognition alone (i.e., the first step only). While providing more comprehensive coverage on the topic of geoparsing, existing efforts reviewed only some approaches for the step of location reference recognition. In recent years, many new approaches for location reference recognition have been developed, such as Flair NER \citep{akbik2019flair}, NeuroTPR \citep{wang2020neurotpr}, nLORE \citep{nLORE}, and GazPNE2 \citep{gazpne2}. Given the high importance of location reference recognition in geoparsing (i.e., only those references that are correctly recognized can be geo-located), it is necessary to have a review that specifically focuses on the possible and recent approaches for location reference recognition.

This work aims at filling the two research gaps discussed above. First, 
we summarize seven typical application domains of geoparsing, which are geographical information retrieval (GIR) \citep{freire2011metadata,purves2018geographic},  %knowledge discovery \citep{tamames2010envmine, weissenbacher2015knowledge, acheson2021extracting},
	 disaster management \citep{lingad2013location,shook2016socio}, %emergency response \citep{cresci2017nowcasting,ozdikis2017survey,de2018taggs,singh2019event}, early warning \citep{avvenuti2016predictability}, 
		 disease surveillance \citep{tateosian2017tracking, gritta2019you,scott2019global},  traffic management \citep{he2013improving,liu2016collective,milusheva2021applying}, spatial humanities \citep{rupp2013customising, gregory2015geoparsing, tateosian2017tracking}, tourism management \citep{colladon2019using,cresci2014towards,brilhante2015planning}, %healthcare accessibility \citep{li2017designing}, 
 and crime management \citep{arulanandam2014extracting,vomfell2018improving,dasgupta2017crimeprofiler}. Second, we review existing approaches for location reference recognition by categorizing the approaches into four groups: rule-based, gazetteer matching-based, statistical learning-based, and hybrid approaches. Noticing that many existing approaches were not cross compared on the same datasets,   %the approach for location reference recognition has not been fully evaluated.,
 we also conduct experiments to compare and evaluate the reviewed \toolss\thinspace existing approaches on 26 public datasets. We examine multiple characteristics of the existing approaches, including their performance on formal and informal texts, their performance on different types of locations (e.g., admin units and traffic ways), and their computational efficiency.

	The remainder of this paper is structured as follows: In Section 2, we summarize seven typical application domains of geoparsing. In Section 3, we review existing approaches for location reference recognition. %We introduce compared approaches in Section 4. 
	We evaluate existing approaches on the same public datasets in Section 4. Finally, we conclude the paper in Section 5 and discuss some potential future directions.

% \usepackage{multirow}

\section{Seven application domains of geoparsing}
\begin{figure}[htbp!]
		\centering
		\includegraphics[width=0.8\textwidth]{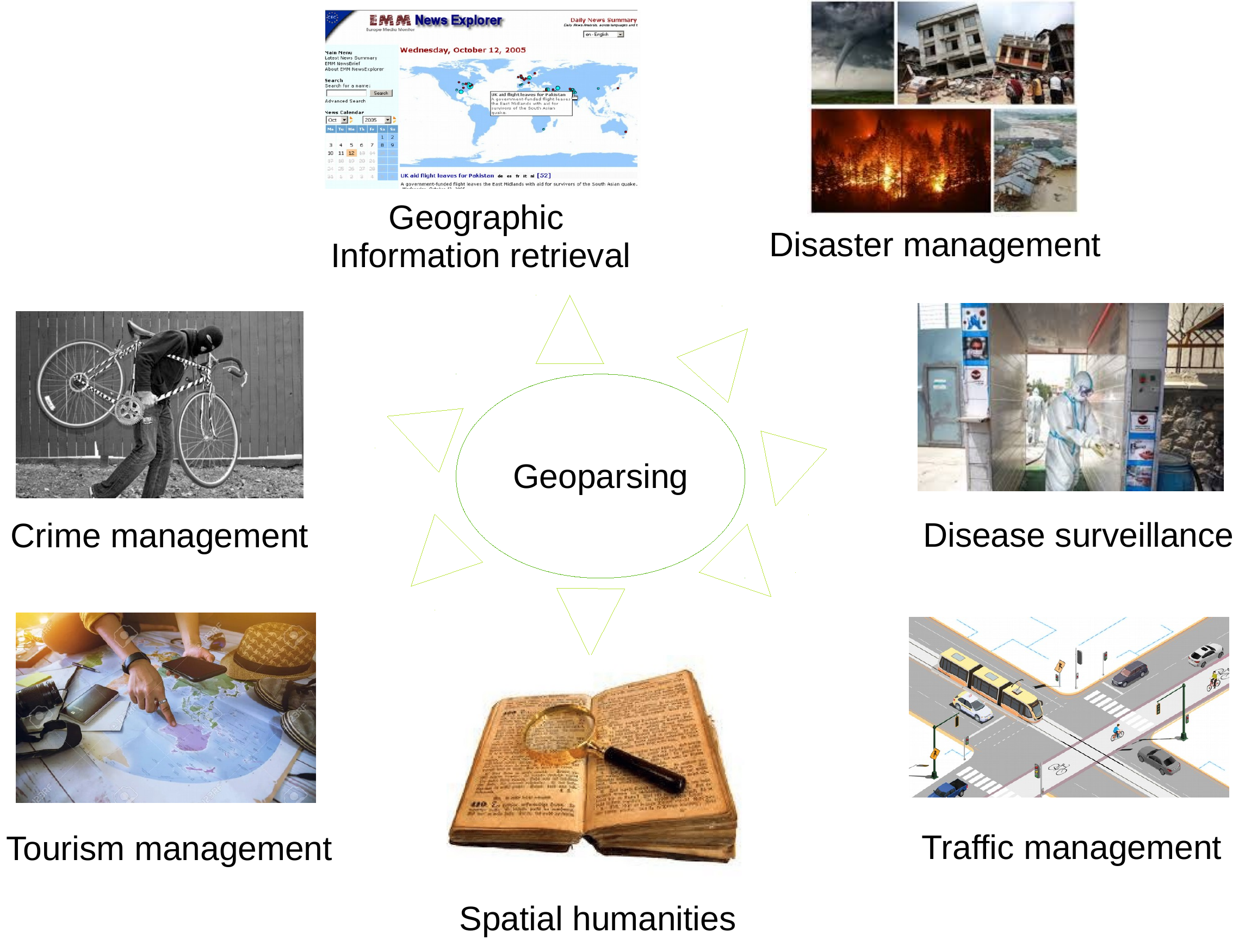}
		\caption{Seven application domains of geoparsing.}
		\label{seven_application_figure}
	\end{figure}

Geoparsing has many possible applications. In this section, we summarize seven typical application domains of geoparsing, which are most discussed in literatures. Figure \ref{seven_application_figure} provides an illustration of these domains.

\textbf{GIR}: One of the primary applications of geoparsing is geographic information retrieval. Historically, documents have been indexed by subject, author, title, and document type. However, a diverse and large group of information system users (e.g., readers, nature resources managers, scientists, historians, journalists, and tourists) desire geographically-oriented access to document collections, such as by retrieving interesting contents about specific geographic locations \citep{woodruff1994gipsy, borges2007discovering, purves2007design, lieberman2012adaptive,freire2011metadata, teitler2008newsstand,mircea2020real}. For instance, resources in digital libraries can be indexed by locations contained in descriptive metadata records associated with the resources, thereby improving users' experience in searching for their needed resources \citep{freire2011metadata}.  People are looking for web pages containing useful information about everyday tasks, such as local merchants, services, and news \citep{borges2007discovering}. The public can consume up-to-date information related to COVID-19 (e.g., disease prevention, disease transmission, and death reports) on Twitter by locations \citep{mircea2020real}.  
 
\textbf{Disaster management}: News stories and SMD contain a large volume of historical and real-time disaster information. Location-enabled SMD can be very helpful to timely map the situational information, such as rescue requests \citep{singh2019event, zou2021social}, resource (e.g., food, clothing, water, medical treatment, and shelter) needs and availability \citep{basu2022utilizing, dutt2019utilizing}, and facility status (e.g., building collapse, road closure, pip broken, and power outage) \citep{fan2020hybrid,scheele2021geographic,belcastro2021using, mao2018mapping} in the aftermath of disasters. With a crisis map, first responders can track the unfolding situation and identify stricken locations that require prioritized intervention \citep{avvenuti2018gsp} and realize optimized real-time resource allocation \citep{singh2019event}, government agencies can conduct the damage assessment of the disasters in a faster manner \citep{wu2018disaster}, and the public can search for the locations where they can obtain needed resources. By extracting spatiotemporal, environmental, and other information about disaster events from news stories,  flood-prone areas can be identified \citep{yagoub2020newspapers}, the responsibility of atmospheric phenomena for floods can be understood \citep{baranowski2020social}, the spatial and temporal distributions of natural disasters during a long period can be analyzed \citep{liu2018analyzing}, %social and economic impact can be automatically summarized, 
and the evolution of disasters (e.g., the phases of preparedness, impact, response, and recovery) can be tracked \citep{wang2015spatiotemporal,huang2015geographic, wang2020tracking}.

\textbf{Disease surveillance:} Scientific articles, historical archives, news reports, and social media contain detailed information of disease events, such as where the disease was first reported and how it spread spatiotemporally. Mining geographic locations and other related information of disease events can help track diseases \citep{tateosian2017tracking, gritta2019you,scott2019global,o2017digital,chunara2012social,murrieta2015automatically}, perform early warning and quick response \cite{keller2009automated}, and understand the mechanisms underlying the emergence of diseases \citep{allen2017global, jones2008global}. %, and predict diseases. 
For example, geoparsing historical archives (e.g., The annual US Patent Office Reports 1840-1850 and Registrar General's Reports) can help track the spread of potato disease ‘late blight’ in the 19th-century in the United States  \citep{tateosian2017tracking} and understand the relationship between cholera related disease and place names during Victorian times \citep{murrieta2015automatically}. %Disease reports were mined and geoparsed to examine the genus-wide distribution and invasiveness of Phytophthora species \citep{scott2019global} and to unveil interesting patterns with regard to the spread and interpretation of the Third Plague Pandemic (1891-1950) \citep{alex2021plague}.
Scientific articles were geoparsed to analyze the demographic, environmental, and biological correlation of the occurrence of emerging infectious diseases at a global scale \citep{jones2008global,allen2017global}. 
%Apart from the location of disease events, the locations and feelings of people during a pandemic are also valuable. For example, extracting patient travel history from unstructured clinical documents is crucial in evaluating evolving infectious disease events, such as to help public authorities work quickly in tracing linkages to manage outbreaks \citep{peterson2021automated} and establish an appropriate differential diagnosis and optimize testing \cite{duong2016importance, alexander2018schistosomiasis}. %Tweets were applied to predict the cholera spread pattern during the 2010 Haiti Earthquake by identifying infected areas mentioned in tweets \citep{chunara2012social}. 
%Moreover, 
%Social media users often share their feelings about diseases online, and 
Social media can also reflect the movement of the public and their feelings during pandemics through geotags or mentioned locations in texts. Location-enabled tweets were applied to analyze the mental health status of the public after the occurrence of COVID-19 \citep{zhu2021using,hohl2021understanding}, to track and visualize the spread and diffusion of COVID-19 \citep{andreadis2021social}, and to reveal human mobility patterns \citep{iranmanesh2021reading, huang2020twitter}.

\textbf{Traffic management:} Twitter users report near-real-time information about traffic events (e.g., crashes and congestion). Detecting traffic events, their precise locations, and other related information from tweets \citep{shang2022sat,giridhar2015quality, gutierrez2015twitter, suat2022extraction,ahmed2019real, alomari2021iktishaf} is important for an effective transportation management system. The detected traffic events can also support urban policy making \citep{das2019exploring}, such as to help drivers to avoid risk zones and choose the fastest and safest routes \citep{ali2021traffic}, to help the transportation
management sector reduce fatalities and restore traffic
flow as  quickly as possible \citep{ali2021traffic}, to predict future traffic jams \citep{alkouz2020snsjam}, and to improve road safety by recognizing high-risk areas \citep{milusheva2021applying}. By doing so, Twitter users acting as social sensors can complement existing physical transport infrastructure (e.g., video cameras and loop detectors) in a cost-effective manner, which is especially important to developing countries where resources are limited. %Table \ref{tweet} lists several tweets, containing precise location information of events.

 \textbf{Spatial humanities:} `Spatial turn' was used to describe a general movement, observed since the end of the 1990s,  emphasizing the
reinsertion of place and space in humanities \citep{warf2008spatial}. Digitizing and geoparsing large historical textual collections, such as books, reports, and novels create new ways for research in humanities (e.g., Archaeology, History, and Literature) \citep{grover2010use, murrieta2015automatically,gregory2016digital, tateosian2017tracking, hinrichs2015trading, donaldson2017locating, moncla2017automated,gregory2015geoparsing}, such as to understand the historical geographies of nineteenth-century Britain and its relationships with the wider world \citep{gregory2016digital},  to identify the significance of specific commodities in relation to particular places and time \citep{hinrichs2015trading}, to analyze a correspondence between eighteenth-century aesthetic theory and the use of the
terms beautiful, picturesque, sublime and majestic in contemporaneous and later accounts of the Lakes region \citep{donaldson2017locating}, and to reveal the spatial structure of a narrative in fictional novels \citep{moncla2017automated}.

% such as (spatial) digital humanities A huge amount of historical books, reports, and archives, including the Old Bailey Online, Early English Books Online, the British Library’s Nineteenth-Century Newspaper collection, and the US Patent Office reports, comprise geospatial information about historical events (such as battles and diseases). Mining the spatial information from historical texts can benefit many applications, such as  \citep{gregory2015geoparsing, won2018ensemble} (e.g., to displaying cultural heritage.) and disease monitoring  (e.g., to track potato disease `late blight' in the 19th-century.)

 \textbf{Tourism management:} According to the prediction of Statista in 2016, there were going to be 32 million active bloggers in the US alone by 2020. Among all the active blogs, travel is rated as the top 5 topics shared by bloggers \footnote{\url{https://www.statista.com/statistics/187267/number-of-bloggers-in-usa/}}. Travel blogs contain a wealth of information about visited places organized as bloggers’ experiences and insights as well as their perceptions of these places \citep{haris2020spatial}. These narratives reflect the blogger's behavior and interaction with places and also the relationships among the places. Geoparsing travel blogs is helpful for understanding places \citep{haris2017mining}, such as to find their features and related activities, and can help describe a place with tourism attributes to support tour planning \citep{kori2007automatic, yuan2016make, haris2017mining, haris2020spatial}. Applications include helping travelers choose preferred places and visit them in an appropriate order at a proper time, and  supporting wayfinding given the spatial relation of places \citep{haris2020spatial}. %, and for the study of urban morphology \citep{crooks2015crowdsourcing}, investigating the physical form and function of urban environments. 

 \textbf{Crime management:} %Law enforcement agencies often have crime information available with them that comes only within their jurisdiction \citep{srinivasa2019crime}.
 Many countries do not make crime data available to their citizens \citep{arulanandam2014extracting} or just provided coarse-grained details \footnote{\url{http://liiofindia.org/}}, such as the total number of thefts in a district or a province. According to the Crime Information Need Survey \citep{arulanandam2014extracting}, around 78.3\% of respondents in Indonesia agreed that crime information should be available to the public. The needed information includes crime type, perpetrator, victim, time, and very importantly, location.
Meanwhile, information related to crime is often scattered across news and social media. Mining and gathering crime-related information from these text-based sources can be useful for informing the public and may even help predict and prevent some crimes  \citep{alruily2014crime,dasgupta2018automatic,sandagiri2020detecting,rahma2021rule,srinivasa2019crime, dasgupta2017crimeprofiler}. In particular, geoparsing can help extract location information of crimes, % such as time, location, type, perpetrator, and victims, from these sources  \citep{}, for law enforcement agencies to analyze and prevent criminal activities \citep{srinivasa2019crime,dasgupta2017crimeprofiler}, 
which can help residents to choose places to live and help travelers to avoid certain unsafe places  \citep{arulanandam2014extracting}. 

Different applications have distinct requirements for the approaches for location reference recognition. For instance, %disease surveillance normally leverages published biomedical articles (formal text), from which the spatiotemporal information of disease events can be extracted, while emergency response normally leverages tweets (informal text); 
informal texts (e.g., tweets) are the main source for emergency response, from which the real-time location of sub events caused by disasters can be extracted, while scientific articles are the main source for analysing the mechanisms underlying the emergence of diseases, from which the spatiotemporal evolution of diseases across the globe can be extracted. GIR just needs coarse-grained geospatial information, such as a city, while traffic management requires the fine-grained location (e.g., a street) of traffic events; Geoparsing historical documents that contain billions of words requires a fast processing workflow. Therefore, to guide the selection of proper approaches for location reference recognition based on application needs, examining the  characteristics of existing approaches is necessary, which will be introduced in Section \ref{method}.

%  Newspaper articles contain valuable information about events (e.g., crimes, disasters, accidents, and politics), such as their location, time, and effect. Harvesting the information from numerous newspapers can draw a more complete picture of the tempo-spatial variations of the events. This is helpful for many important issues, such as to prevent theft crime \citep{arulanandam2014extracting}, to discover flood-prone areas , and to benefit political science study \citep{gunasekaran2018sperg}. %Besides, newspapers tend to characterize their readership in terms of location, and publish news articles describing events that are relevant to geographic locations of interest to their readers \citep{lieberman2011multifaceted,lieberman2012adaptive}. 
 
% 	Scientific articles often contain relevant geographic information \citep{tamames2010envmine, leveling2015tagging, weissenbacher2015knowledge, magge2018deep,davari2019toponym,  acheson2021extracting,palmblad2017spatiotemporal,allen2017global,jones2008global}, such as where is an author's affiliation, where fieldwork was performed, where were sampling sites, where patients were treated, where an interview was conducted, and where a virus or disease was reported. Automatically extracting the geographic information could help conduct meta-analyses (e.g., to detect and monitor emerging infectious diseases \citep{allen2017global,jones2008global}) and find geographical research gaps (e.g., to perform spatiotemporal bibliometric analyses of research in tropical medicine \citep{palmblad2017spatiotemporal}).  

	\section{A survey of existing approaches}
	
	In this section, we review existing approaches for location reference recognition. In subsection 3.1, we review individual approaches by categorizing them into four groups, and in subsection 3.2, we review existing comparative studies and differentiate our current review from the existing studies.%  and comparative and survey studies related to geoparsing.

	\subsection{Approaches for location reference recognition}
	
\begin{figure}[htbp!]
		\centering
		\includegraphics[width=0.6\textwidth]{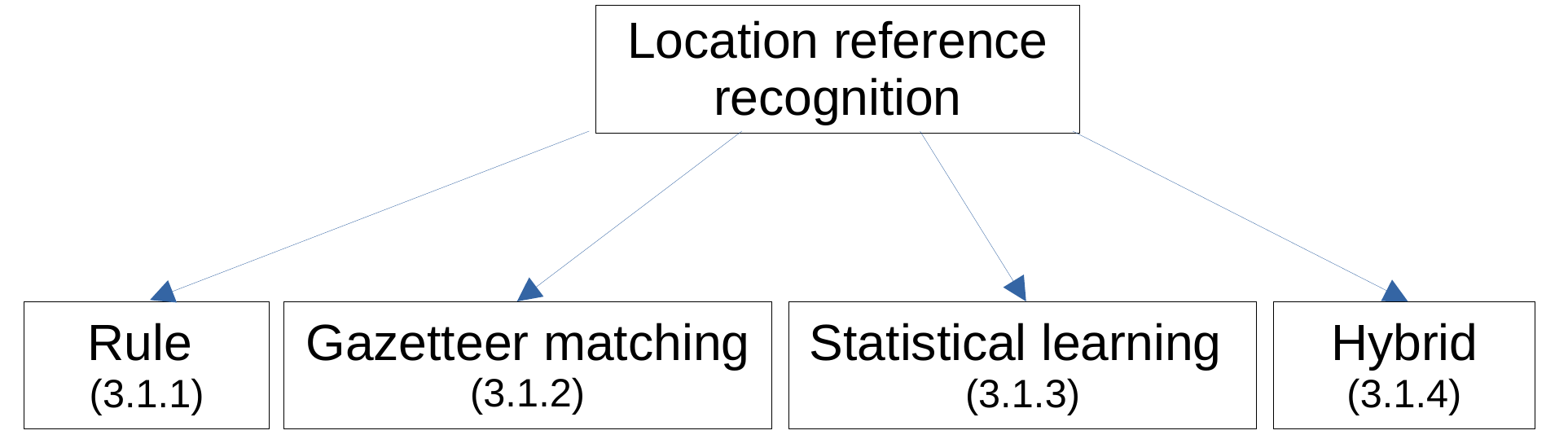}
		\caption{Classification of existing approaches for location reference recognition.}
		\label{class}
	\end{figure}

	 In the existing literature, \citet{leidner2011detecting,monteiro2016survey,purves2018geographic} identified three types of approaches for location reference recognition, which are rule-based, gazetteer matching-based, and statistical learning-based. However, many studies, such as  \citep{gelernter2013algorithm,li2014fine,martinez2020knowledge,hu2020GazPNE}, used a combination of different approaches to compensate the shortcomings of each other. Therefore, in this review, we add a fourth type, \textit{hybrid approaches}, which combine two or all three types of approaches, and we use these four types to organize our review on location reference recognition. We show this classification schema  in Figure \ref{class}. %(1) Rule-based approaches, (2) Gazetteer-matching-based approaches, (3) Statistical learning-based approaches, (4) Hybrid approaches that combine two or all of three approaches. %However, in the following, we do not discuss rules-based approaches with an independent subsection  since pure-rules based approaches are rare. Instead, we introduce the rule-based approaches together with hybrid approaches, since rules are nominally combined with gazetteers. 
	
	%Before the social media era, researchers focused on extracting locations from online contents such as web pages, news, and blogs.
	\subsubsection{Rule-based approaches}
	%Place names in texts have certain lexical and contextual characteristics. Linguistic patterns, 
Location references in texts often have certain lexical, syntactic, and orthographic features. %, and POS features. 
A set of rules, such as regular expressions (REs) and context-free grammars (CFGs), can be defined to decide if an n-gram of texts is a location reference or not \citep{leidner2011detecting}. n-grams are linear sequences of $n$ words in texts. For example, given a text $T=\{w_0 \, w_1 \, w_2...w_n\}$, its unigrams or n-grams of size $n=1$ include $\{w_0\}$,$\{w_1\}$,$\{w_2\}$...,$\{w_n\}$. Its bigrams or n-grams of size $n=2$ include $\{w_0 \, w_1\}$,$\{w_1 \, w_2\}$...,$\{w_{n-1} \, w_n\}$.

Table \ref{res} lists some RE and grammar rules  used in previous studies \citep{leidner2011detecting, gelernter2013cross,giridhar2015quality,martinez2020knowledge}. Each row in the table indicates a rule. The former twelve rules are REs, using Part-of-speech (POS) tags and/or keywords. We use the standard meta characters (i.e., `$?$', `$+$',  and `$*$') of REs. The `$?$' sign indicates the presence of a tag zero or one time, the `$+$' sign indicates the presence of a tag at least one time, and the `$*$' sign indicates the presence of a tag any times (zero or more).  %the `$|$' sign indicates the presence of one of the two tags. 
 Numbers indicate different types of words. 1 represents street indicators, such as \textit{'street'}, \textit{'highway'}, \textit{'road'}, \textit{'sh'}, and \textit{'beltway'}. 2 represents words that specify direction or a distance in measurable terms, such as \textit{`10'}, \textit{`away'}, \textit{`from'}, \textit{`miles'}, \textit{`km'}, \textit{`south'},  and \textit{`northbound'}. 3 represents place  category words, such as \textit{`city'}, \textit{`str'}, \textit{`avenue'}, \textit{`rd'}, and \textit{`village'}.  Used POS tags include Nouns (NN), Proper Nouns (NNP), Determiners (DT), Adjectives (JJ), Cardinal Numbers (CD), and Conjunctions (CC). The last two rules are grammar rules. $X$ denotes candidate n-grams. 4 represents place category words that are often used with \textit{`of'}, such as \textit{`city', `town', `gulf',} and \textit{`river'}. 5 represents spatial prepositions that normally appear before a location, such as \textit{`in'}, \textit{`around'}, \textit{`on'}, \textit{`near'}, and \textit{`between'}. %Note that there might be conflicts among these rules since they were created in different studies.

  \begin{table}[htp!] \centering  \footnotesize  \caption{Examples of RE and grammar rules for location reference recognition.} \label{res}
 
\begin{tabular}{|c|c|}
\hline
Regular Expression                                                                                                 & Examples                                                         \\ \hline
\textless{}NN\textgreater{}+                                                                                     & tiburon blvd; san manteo                                         \\ \hline
\textless{}NNP\textgreater{}+                                                                                    & Heidelberg; San Francisco                                        \\ \hline
\textless{}DT\textgreater{}?\textless{}JJ\textgreater{}?\textless{}NN\textgreater{}+                             & the golden gate bridge; long island                              \\ \hline
\textless{}CD\textgreater{}?\textless{}DT\textgreater{}?\textless{}JJ\textgreater{}?\textless{}NN\textgreater{}+ & third street; 11th avenue                                        \\ \hline
\textless{}DT\textgreater{}?\textless{}JJ\textgreater{}?\textless{}NN\textgreater{}+\textless{}CD\textgreater{}? & freeway 91; highway 12                                           \\ \hline
\textless{}DT\textgreater{}?\textless{}JJ\textgreater{}\textless{}NN\textgreater{}(1)                            & the high cotton lane; high star drive                            \\ \hline
(1)\textless{}CD\textgreater{}                                                                                   & beltway 10; sh 73                                                \\ \hline
(2)+\textless{}NNP\textgreater{}+(3)*                                                                            & south Northumbria bridge road; northeast Munich                  \\ \hline
\textless{}NNP\textgreater{}+(3)*(2)+                                                                            & Camanche Avenue east; Heidelberg North                           \\ \hline
(2)+(3)*(of)?\textless{}NNP\textgreater{}+                                                                       & 25 miles sw of San Francisco; 25min away from New York State     \\ \hline
(3)*(of)?\textless{}NNP\textgreater{}+(2)+                                                                       & town of San Francisco; district of Columbia                      \\ \hline
\textless{}A-Z\textgreater{}\textless{}a-z\textgreater{}*berg                                                    & Heidelberg; Freiberg                                             \\ \hline
(4) (of) X -\textgreater \textless{}LOC\textgreater{}                                                            & city of beaumont; Gulf of Mexico;                                \\ \hline

(5) X -\textgreater \textless{}LOC\textgreater{}                                                                 & this overturned tanker in \textbf{marin} has created a huge jam on \textbf{wb 580} \\ \hline

\end{tabular}
\end{table}

There are several studies that used only rules to extract location references. For instance, \citet{giridhar2015quality} used road-traffic-related tweets to detect and locate point events, such as car accidents. Specifically, a set of REs were defined according to the composition of nouns, determiners, adjectives, cardinal numbers, conjunctions, and possessive endings. Furthermore, to decrease false positives, grammar-based rules were implemented based on spatial prepositions, such as \textit{in}, \textit{at}, \textit{between}, and \textit{near}. %Geocoding is conducted with Google Map API.
\citet{zou2021social} analyzed the rescue request on Twitter during Hurricane Harvey. The authors first manually annotated the tweets that were asking for help. Then, they used a rule-based method to recognize location references from tweets. %, which were then geocoded with Google Map API. 
More specifically, they assumed that the formal description of an address in the United States is in the form of [Street Number, Street Name, Apartment Number (optional), City, State, Zip Code]. Since all rescue request tweets in their study contain zip codes, the full address in each tweet can be extracted by locating the zip code as the ending point and searching for the starting point based on several conditional criteria. %For example, if there is only one number before the zip code, then the full address is the text from the first number to the zip code.

 %Therefore, there are few works that use pure rule-based systems to extract place names from texts. Nevertheless, they are still useful since they can be combined with gazetteers or statistical learning to compensate for each other.	
Although many studies classify rule-based approaches as one category \citep{leidner2011detecting, monteiro2016survey, al2017location}, pure rules-based approaches are rare. All the rule-based approaches discussed in \citep{monteiro2016survey} are, in fact, hybrid approaches. This is likely because the approaches that rely  on linguistic patterns only are ineffective \citep{silva2006adding}. It remains a challenge to  define rules in a complete and robust manner that can explain all possible occurrences of location references in texts, especially in microblogs with dramatic variation of writing styles and weak grammar \citep{ritter2011named}. However, a set of simple rules can be used to enhance gazetteer matching and statistical learning-based approaches, which will be introduced in the following subsections.
	
%  \citet{bontcheva2013twitie} presented TwitIE-GATE, an open-source NLP pipeline customized to microblog text at every stage, which adapts rules from ANNIE \citep{cunningham2002gate} for the extraction of name entities. ANNIE consists of the following main processing resources: a tokenizer, sentence splitter, POS tagger, gazetteer lists, finite state transducer, orthomatcher and coreference resolver.

 %TwitIE-GATE is evaluated on a corpus of 2,400 tweets comprising 34,000 tokens. Finally, an F1 score of 0.80 was achieved on the place name recognition task.  
 
   % Generally, the rule-based approach is highly efficient in computation. In some cases, it can achieve a promising tagging result. However, it remains challenging to define complete and robust rules that can adapt to all the variations of place names in texts, especially in microblogs, considering the dramatic variation of the writing styles, weak grammar rules, and numerous grammar errors in microblog texts \citep{ritter2011named}. Therefore, there are few works that use pure rule-based systems to extract place names from texts. Nevertheless, they are still useful since they can be combined with gazetteers or statistical learning to compensate for each other.

\subsubsection{Gazetteer matching-based approaches}

% \usepackage{multirow}

\begin{figure}[htbp!]
		\centering
		\includegraphics[width=0.8\textwidth]{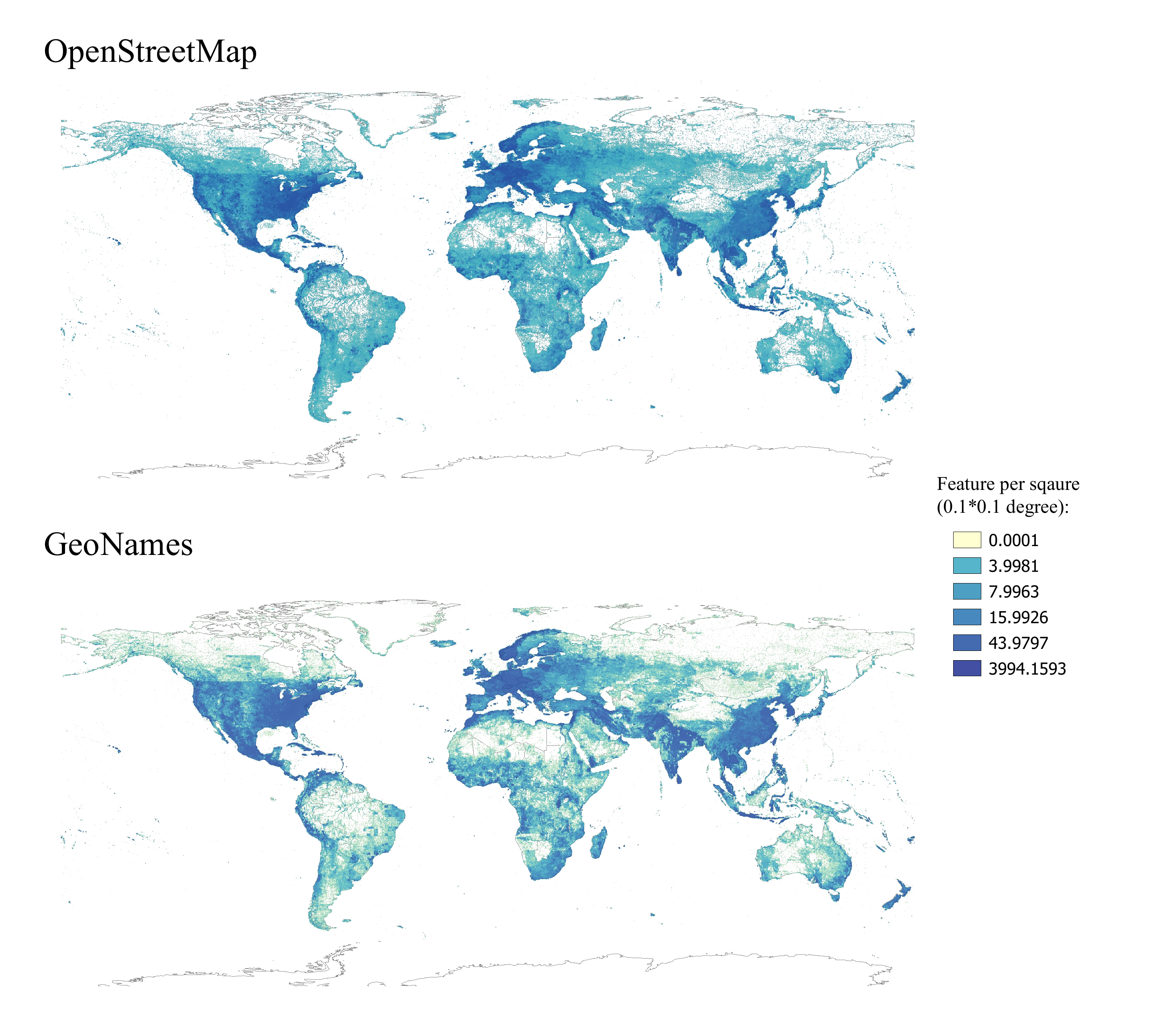}
		\caption{Point density maps of the places contained in OpenStreetMap and GeoNames.}
		\label{density}
	\end{figure}

A gazetteer is a dictionary of place names %or geographical thesaurus 
associated with geospatial information (e.g., place types and geographic coordinates) and some additional information such as population size, administrative level, and alternative names. %Most existing studies used gazetteers for the toponym resolution step (i.e., the 2nd step of geoparsing) in spite of the existence of some gazetteer-free geocoding approaches \citep{kulkarni2021multi, delozier2015gazetteer}. Moreover, 
Gazetteers play important roles in location reference recognition in many studies. GeoNames \footnote{\url{http://www.geonames.org/}} is a most widely used gazetteer, and  OpenStreetMap (OSM) \footnote{\url{https://www.openstreetmap.org/}}, in a broad sense, can be considered as a gazetteer as well. %, and GNIS/GNS. 
%Gazetteers vary in the coverage of places and associated information. 
There are 12,255,028 \footnote{Retrieved from the official web of GeoNames on 2022.02.25} and 23,876,956 \footnote{Retrieved from OSMNames on 2022.02.25}   places in  GeoNames and OSM, respectively. Figure \ref{density} illustrates the point density map of the places in OSM and GeoNames.

%Gazetteers were also described with  terms, such as knowledge base, ontology, database, or dictionary. In this study, they are named gazetteers. 
In gazetteer matching-based approaches, the n-grams of a text are first matched against a gazetteer, which are then filtered or disambiguated with a couple of heuristics. %This can not only recognize place names but also specify their geo-coordinates. 
Gazetteer matching-based approaches are still faced with two main challenges. The first is that many location references appearing in texts are missing from gazetteers due to various reasons, such as name variation (e.g., \textit{`South rd'} for \textit{`South road'} and \textit{`Frankfurt airport'} for \textit{`Frankfurt international airport'}) and data incompleteness (e.g., the missing of \textit{`Hidden Valley Church of Christ'} from a gazetteer) \citep{gelernter2013algorithm}. Second, gazetteer matching-based approaches often run into ambiguity issues. For instance, the names \textit{`Washington'},   \textit{`MO'}, \textit{`South Wind'}, and \textit{`1 ft'} all exist in gazetteers, but can also refer to other types of entities. This is called geo/non-geo ambiguities, while geo/geo ambiguities refer to the situation that different spatial locations use the same name, such as Manchester, NH, USA versus Manchester, UK. For simplicity, we use \textit{ambiguities} and \textit{ambiguous} to  refer to geo/non-geo ambiguities by default. We will use the full name \textit{geo/geo ambiguities} to refer to the second situation. 
The main focus of gazetteer-based approaches is often to overcome the two mentioned challenges by using heuristics to perform disambiguation (to increase precision)  and by including  place name variants to expand the used gazetteer (to increase recall).

 Table \ref{heuristics} summarizes the commonly used heuristics for disambiguation. The first four heuristics are used to reduce the number of candidate place names matched in gazetteers, thereby decreasing the number of ambiguous place names. The 5th heuristic uses common words. The 6th and 7th heuristics leverage the external and internal cues of candidate n-grams. The 8th heuristic leverages the POS tags of candidate n-grams. The 9th heuristic leverages the dictionary of other entity types (e.g., \textit{Person}), such as to judge if a candidate n-gram (\textit{`Houston'}) with its preceding or succeeding word (\textit{`Alexander'}) in texts appears in the dictionary of person names. The 10th heuristic leverages other related place names (e.g., nearby or with the same name) to judge if an n-gram is valid or not. For example, \textit{`IN'} is ambiguous. However, when it precedes \textit{`Chennai'} in texts (e.g., \textit{`Chennai, IN'}), which is likely to be a location and related to \textit{`IN'}, then \textit{`IN'} is treated as a valid location. If an n-gram can be determined as a valid location by some heuristics, the other n-grams with the same name in the text are also treated as a valid location, such as  \textit{`stay safe Houston, flood in Houston is serious'}, where both \textit{`Houston'} are judged as valid locations since the preceding word (\textit{`in'}) of the second \textit{`Houston'} is a spatial indicator.

{\begin{table}[htpb!] \footnotesize \caption{Common heuristics used for disambiguation in gazetteer matching-based approaches}  \label{heuristics}
\begin{tabular}{|l|c|c|}
\hline
ID & Heuristics                                & Examples                                                           \\ \hline
1  & Limit the length  of places in gazetteers & Keep only 1- and 2-grams                                           \\ \hline
2  & Limit the type of places in gazetteers    & Keep only continent, country, state, and city                      \\ \hline
3  & Limit the  scale of places in gazetteers  & Keep only places with a population over 1000                       \\ \hline
4  & Limit the spatial range of gazetteers     & Use the gazetteers in the area of Florence                         \\ \hline
5  & Filter place names of common (stop) words & `today', 'long', `that building' and `the street'                  \\ \hline
6  & Use spatial indicators in texts           & `in', `near', and `at' that appear before a place                  \\ \hline
7  & Use orthographic cues                     & Capitalization of words, such as `Houston' and  `Germany'          \\ \hline
8  & Filter candidates by POS tags              & Keep only noun phrases in texts                                     \\ \hline
9  & Use a dictionary of other entity types      & Person names, such as`Washington Irving' and `Houston Alexander'   \\ \hline
10 & Use other related place names             & `Chennai, IN' and `stay safe Houston, flood in Houston is serious' \\ \hline
\end{tabular}
\end{table}
}

Many studies used gazetteer matching-based approaches to recognize location references from texts \citep{woodruff1994gipsy,amitay2004web,clough2005extracting,sultanik2012rapid,paradesi2011geotagging,middleton2018location,pouliquen2006geocoding,silva2006adding, tamames2010envmine,allen2017global,paradesi2011geotagging,de2018taggs,ferragina2010tagme,al2017location,ahmed2019real,milusheva2021applying}. %Table \ref{gaz} summarizes representative approaches for location reference recognition based on gazetteer matching. The column `Heu' lists the IDs of used heuristics in Table \ref{heuristics}. 
One of the earliest geoparsing approaches was proposed by \citet{woodruff1994gipsy} to support georeferenced document indexing and retrieval. A gazetteer containing around 120,000 places in California is first built on the US Geological Survey's Geographic Names Information System (USGS 1985) and the land use data from the US Geological Survey's Geographic Information Retrieval and Analysis System (GIRAS). The place names in a document were identified by matching texts' n-grams containing non-stop words against the gazetteer. If a token had no matches in the gazetteer, it was depluralized (e.g., \textit{`valleys'} to \textit{`valley'}) and rematched with the gazetteer. Identified place names were then geocoded  to determine the geographic scope of the document. \citet{amitay2004web} developed a system, named Web-a-Where, for recognizing and geocoding continents, countries, states, and cites as well as their abbreviations in web pages.  A gazetteer was created by collecting about 75,000 place names across the world from different data sources: USGS, World-gazetteer.com \footnote{\hyperlink{http://www.world-gazetteer.com}{http://www.world-gazetteer.com}}, UNSD \footnote{\hyperlink{http://unstats.un.org/unsd}{http://unstats.un.org/unsd}}, and ISO 3166-1 \footnote{\hyperlink{https://www.iso.org/iso-3166-country-codes.html}{https://www.iso.org/iso-3166-country-codes.html}}. The system first extracted candidate place names in a given page by matching against the gazetteer. Then, four heuristics were sequentially used to disambiguate and geocode the candidate place name, such as the vicinity of two candidate places (e.g., Chicago, IL) and the population of places. %, multiple candidate places of the same name in a page, and the smallest disambiguating context of two candidate places. 
The approach was evaluated on 600 web pages, containing over 7000 place names. 
\citet{clough2005extracting} proposed geoparsing web pages. In the recognizing step, candidate place names are first identified by matching against gazetteers, which are then filtered by using stop words and by using context cues, such as to filter person names with simple heuristic $<title><loc> $ (e.g., Mr. Sheffield), where $<loc>$ is a candidate place name and also in the dictionary of person names. Used gazetteers include the Ordnance Survey 1:50,000 Scale Gazetteer for the UK (OS \footnote{\hyperlink{http://www.ordnancesurvey.co.uk/oswebsite/products/50kgazetteer/}{http://www.ordnancesurvey.co.uk/oswebsite/products/50kgazetteer/}}), Seamless Administrative Boundaries
of Europe dataset (SABE \footnote{\hyperlink{http://www.eurogeographics.org/eng/03\_projects\_sabe.asp}{http://www.eurogeographics.org/eng/03\_projects\_sabe.asp}}), and Getty Thesaurus of Geographic
Names (TGN \footnote{\hyperlink{http://www.getty.edu/research/conducting\_research/vocabularies/tgn/}{http://www.getty.edu/research/conducting\_research/vocabularies/tgn/}}). %In the geocoding step, several heuristics (e.g., the admin level of places) are used to deal with the geo/geo ambiguities. 130 web pages with 1864 places were used as the test data. %, and an F1 score of 0.71 is achieved in location reference recognition. 
%The proposed geoparsing approach was utilized in SPIRIT \citep{purves2007design}, a Geographic Information Retrieval system for the web.
\citet{pouliquen2006geocoding} proposed  geoparsing multilingual texts. Candidate place names are first identified by matching with a multilingual gazetteer, which are then disambiguated through a dictionary of person names (e.g., \textit{`George Bush'} and \textit{`Tony Blair'}) and stop words (e.g., \textit{`And'}, \textit{`Du'}, \textit{`Auch'}) in a multilingual context. %In the geocoding step, geo/geo ambiguities are removed by combining several heuristics, such as the importance of places, the main countries a text is about, and the minimum physical distance between ambiguous places and unambiguous ones. 
The multilingual gazetteer is created from three sources: Global Discovery database of place names (Global Discovery 2006), the multilingual KNAB database (KNAB 2006), and a European commission internal document. The approach was tested on 162 newspaper stories in five languages (i.e., German, English, Spanish, French, and Italian) from Europe Media Monitor. %to evaluate the proposed approach and an F1 score of 0.77 was achieved.

Gazetteer matching-based approaches were also used to extract locations from tweets. For instance, \citet{paradesi2011geotagging} proposed TwitterTagger for geoparsing tweets. It matched the noun phrases of a tweet text with the entries in gazetteers (i.e., USGS database), which was followed by disambiguating the matched entry with two heuristics. The first was to check if spatial indicators (e.g., \textit{`in'} and \textit{`near'}) were used before a noun phrase. The second was to check whether other users used a spatial indicator before the same noun phrase in their tweets. Geo/geo ambiguities were removed by calculating the distance between the location of a place name in a tweet and the location of the user who posted the tweet as well as the location of other tweet users who mentioned the same place name. The approach was evaluated on 2000 annotated tweets. % \citet{sultanik2012rapid} proposed RapidGeo, which can perform a fuzzy matching between strings on a K-D tree structure, mapping toponyms to likely locations in  GeoNames. The fuzziness enables mitigating misspellings, such as \textit{`Columbia'} vs. \textit{`Colombia'}, in the input data. The approach was evaluated on 500 tweets.
\citet{middleton2018location} proposed a multilingual geoparser for tweets, named \textit{geoparsepy}. To overcome the place name variation issues, a set of heuristics were applied to expand OSM place names. To deal with abbreviations, a multilingual corpus of the street and building types from OSM was used to compute obvious variants for common location types (e.g., \textit{`Southampton Uni'} for \textit{`Southampton University'}). To overcome the ambiguity issue, uni-gram location names that are non-nouns are filtered using a multilingual WordNet corpus lookup, such as \textit{`ok'} and \textit{`us'}, which can refer to locations or other types depending on their POS tag. Location phrases are then filtered using a multilingual stop word corpus. To remove the geo/geo ambiguities, a confidence score is calculated for each matched entry in the gazetteer based on several evidential features, such as the other location references in the tweet, the admin level, and the geotag of the tweet if available. %They evaluated the approach on four datasets, corresponding to the earthquake in New Zealand in 2011, the hurricane in New York in 2012, the Blackout in Italy in 2013, and the earthquake in Turkey in 2012, consisting of 2000 (English), 1996 (English), 391 (Italian), and 2000 (Turkish) tweets, respectively. 
\citet{de2018taggs} proposed a geoparsing algorithm (named TAGGS) by employing both metadata and the contextual spatial information of groups of tweets referencing the same location regarding a specific disaster type. It matches the uni- and bi-grams of a tweet text with the GeoNames gazetteer. The found candidates are then filtered with several heuristics, such as discarding the candidates with the 1000 most common words. %To remove geo/geo ambiguities, several spatial evidence (UTC offset, User hometown, geotag, and other related places mentioned in the same tweet) were utilized to score a matched place in the gazetteer. Furthermore, the authors assumed that multiple tweets that mentioned the same toponym within a given timeframe referred to the same location. Based on this assumption, the score of a matched place is further refined by grouping the tweets.    %, and consider small towns, with a population of at least 1, if the town mentioned in the text is written with a capital letter. 
%The approach was evaluated on 2785 flood-related tweets. % and an F1 score of 0.87 is achieved in location reference recognition. 
%TAGGS was used by \citet{baranowski2020social} to geoparse the tweets related to floods  to support the understanding of atmospheric conditions leading to floods in Sumatra. 
Some studies \citep{avvenuti2018gsp, nizzoli2020geo} leveraged gazetteer matching-based semantic annotators to realize their own geoparsing tool. For instance, \citet{nizzoli2020geo} employed TagMe \citep{ferragina2010tagme} to identify entities of an input text and to link them to the corresponding entity in a knowledge graph (i.e., DBpedia). Then, they traversed the knowledge graph to expand the available information for the geoparsing task. Finally, they exploited all available information for learning a regression model that selects the best entity in the knowledge graph for annotated places in the text. %The proposed geoparsing approach was evaluated on two twitter datasets. The first is GeoCorpora that is a public tweet data set for the evaluation of geoparsing approaches. The second is NEEL2016 \citep{rizzo2015making} that is the official dataset of the 2016 Named entity recognition and linking challenge.

Studies, such as \citep{al2017location, ahmed2019real,xu2019dlocrl, yagoub2020newspapers, milusheva2021applying,belcastro2021using,suat2022extraction}, focused only on local events whose geographical scope is known, such as floods or traffic accidents happened in a certain city. Therefore, they would normally use a local gazetteer that contains only the places in a certain region, which can dramatically mitigate the issues of geo/non-geo ambiguities and geo/geo ambiguities. Although the proposed geoparsing approaches are not  globally applicable, they are effective in dealing with local events. For instance, \citet{al2017location} proposed a Location Name Extraction tool (LNEx), which used n-gram statistics and location-related dictionaries to handle the abbreviations and automatically filter and augment the place names in the OpenStreetMap gazetteer (handling name contractions and auxiliary contents). %LNEx was evaluated on 4500 event-specific tweets, related to three foods: 2016 in Louisiana (US), 2016 in Houston (US), and 2015 in Chennai (India). %An F1 score of 0.81 was achieved by LNEx. 
%When configuring LNEx for each test scene, only the OpenStreetMap data in the affected region (e.g., Houston) was utilized. 
\citet{ahmed2019real} used tweets to monitor the traffic congestion in real-time. Specifically, tweets related to traffic congestion are first detected by using supervised and unsupervised machine learning techniques, and the road names in the tweets are then extracted by matching with a list of road names in the city of Chennai. Jaro-Winkler metric is used to calculate the similarity between the n-grams in tweets and the road names in the list to overcome the challenge of place name variants. 1551 congestion-related tweets were used to detect the congestion situations in Chennai in two short periods, lasting 7 months in total. 
\citet{milusheva2021applying}  used traffic-related tweets to derive the locations of road traffic crash in Nairobi, Kenya, for the purpose of road safety improvements. Specifically, they applied a machine learning model to capture the occurrence of a crash and developed a gazetteer matching-based geoparsing algorithm to identify its location. A gazetteer of landmarks (e.g., roads, schools, and bus stops) for five counties that constitute the Nairobi metro area was created from OpenStreetMap, GeoNames, and Google Places. The location of crashes is then determined by matching the n-grams of the tweets with the entries in the gazetteer.  %Recently, \citet{belcastro2021using} presented a new method that analyzes tweets to discover sub-events that occurred after a disaster, such as collapsed buildings, broken gas pipes, and flooded roads. Recognizing and geocoding the location mentions in tweets is one of the key tasks. Specifically, a local gazetteer is first created by retrieving points of interest (POIs) in a disaster area from geocoding web services. Then, street and district names from a tweet text are extracted and then geocoded through gazetteer matching. The geocoded coordinates contain four levels of accuracy: POI, street, district, and city. 

It is simple to implement gazetteer matching-based approaches and they can easily adapt to a multilingual context. Moreover, they are effective in certain applications, such as the ones whose geographic scope is limited to a small region (a city) or the ones require only coarse-grained locations, such as country names. However, it remains a challenge to propose a general and globally applicable approach for location reference recognition by using gazetteer matching and simple heuristics since the name variants and geo/non-geo ambiguity issues are ubiquitous in natural language texts. To overcome this challenge, many studies combined gazetteer matching with rules and/or statistical learning to compensating the shortcomings of each other, which will be introduced in the following subsections.

	\subsubsection{Statistical learning-based approaches}

\begin{figure}[htbp!]
		\centering
		\includegraphics[width=0.5\textwidth]{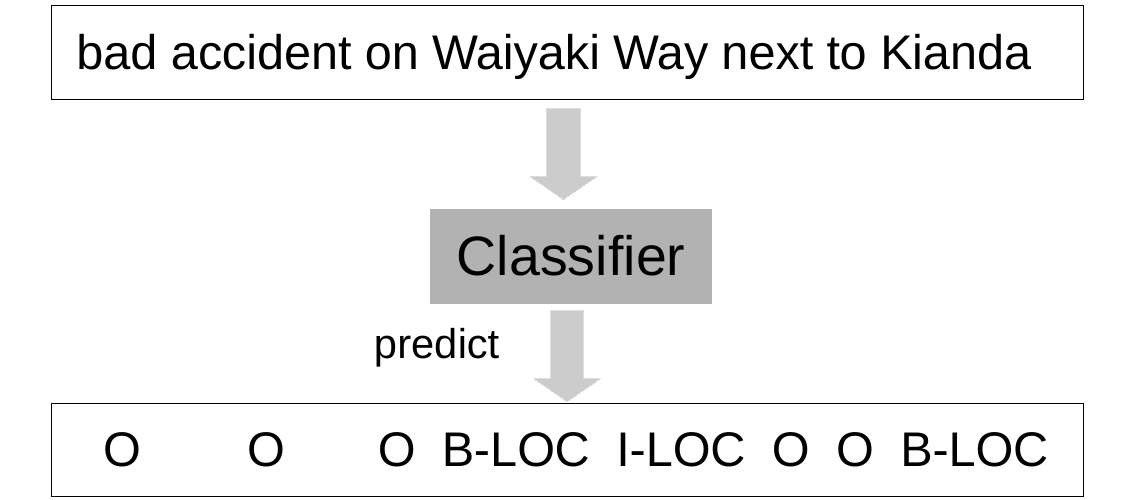}
		\caption{Basic architecture of statistical learning-based approaches. \textit{O} denotes non-type. \textit{B-LOC} and \textit{I-LOC} denote the beginning and inner part of a location reference, respectively.}
		\label{learning}
	\end{figure}

Statistical learning-based approaches are built on annotated training corpora containing texts associated with the expected location references. The annotated corpora are used to train a model via manually defined features, such as infrequent strings, length, capitalization, and contextual features, and/or features automatically learned by deep learning algorithms. The trained model is then applied to unlabeled texts, and the same features are computed to decide on the association of texts and location references. The basic architecture of  statistical learning-based approaches is illustrated in Figure \ref{learning}. 
Statistical learning-based approaches generally use either traditional machine learning models, such as {Random Forest (RF) \citep{freire2011metadata}, %Conditional Random Fields (CRF) \citep{inkpen2017location}, and Maximum Entropy (ME) \citep{nissim2004recognising}}  
or deep learning models, such as %Convolutional Neural Network (CNN) \citep{kumar2019location}, 
Long Short-Term Memory (LSTM) \citep{wang2020neurotpr}. % and Transformer \citep{khanal2021multi}. 
 %Table \ref{sta} summarizes representative approaches for place name recognition based on statistical learning. 
 Statistical learning-based approaches can be further divided into two groups: \textit{learning-based 
 named entity recognition (NER)} and \textit{learning-based place name extraction (PNE)}. In the following, we discuss these two groups of approaches respectively.

 \textbf{{Learning-based NER}}: Location reference recognition can be considered as a subtask of NER, which has been extensively studied. Therefore, many studies \citep{gelernter2011geo, lingad2013location,unankard2015emerging,halterman2017mordecai,mao2018mapping, kamalloo2018coherent, gutierrez2015twitter} used existing statistical-based NER models or retrain them to extract location references from texts. For instance, \citet{lingad2013location} used OpenNLP \footnote{ \url{https://opennlp.apache.org/}}, TwitterNLP  \citep{ritter2011named}, Yahoo!Placemaker, and Stanford NER to extract place names from 2878 disaster-related tweets. Stanford NER and OpenNLP were also retrained and evaluated by using 10-fold cross-validation in their study. The results show retrained models achieved a much higher F1 score than pretrained models. %, with Stanford NER achieving the highest F1 score of 0.9. 
 \citet{karimzadeh2019geotxt} proposed a geoparsing system for tweets, named GeoTxt. It combined six publicly available NERs for location reference recognition, which are Stanford NER, Illinois CogComp \citep{ratinov2009design}, GATE ANNIE \citep{bontcheva2013twitie}, MIT IE \footnote{\url{https://github.com/mit-nip/MITIE}}, Apache OpenNLP, and LingPipe \footnote{\url{http://alias-i.com/lingpipe/demos/tutorial/ne/read-me.html}}. %Several heuristics involving  other places in the same text were used to remove the geo/geo ambiguities in geocoding, such as to minimize the hierarchical relationship between the places. GeoTxt was evaluated on GeoCorpora. %Results show CogComp achieves the highest F1 score of 0.78. 
 \citet{belcastro2021using} utilized  tweets to discover sub-events that occurred after a disaster, such as collapsed buildings, broken
gas pipes, and flooded roads. Geoparsing tweets is one of the key tasks. Specifically, CoreNLP \citep{manning2014stanford} is adopted to recognize street and district names, which are then geocoded by matching with a local gazetteer, covering the disaster area. \citet{fan2020hybrid} proposed uncovering the unfolding of disaster events based on tweets. Situational information (e.g., affected individuals, injured people, recuse, and damage) were extracted by using Stanford NER, which were then filtered and geocoded by keeping only the matched places in Google Geocoding API and by excluding the places outside affected areas. To understand the geographic origin and
patterns of spread of the potato disease `late blight' for the 19th-century US and European outbreaks and the means by which it spread, \citet{tateosian2017tracking} used CLAVIN \footnote{\url{https://github.com/Novetta/CLAVIN}} to geoparse two historical collections: US Patent Office Reports 1841-1850 and Google Books Corpus. CLAVIN is an open-sourced geoparser that utilizes Apache OpenNLP for place name extraction. \citet{mircea2020real} implemented a prototype dashboard for real-time classification %(e.g., disease transmission, disease treatment, and death reports)
, geolocation, and interactive visualization of COVID-19 tweets. spaCy \footnote{\hyperlink{https://spacy.io/}{https://spacy.io/}} was used to extract cities and countries from tweet content and user profiles. %They were then matched with GeoNames to determine their coordinates. %Possible geo-tags of tweets were also provided as another type of location in the system.
 \citet{mao2018mapping} proposed mapping near-real-time power outages from tweets using a retrained NeuroNER model \citep{dernoncourt2017neuroner}. %that adopts a BiLSTM-CRF architecture using both word and char-based embeddings was retrained to extract location references from the power outage tweets. Last, the location references were geocoded by using Bing Maps Location API. The gold data of ALTA 2014 Twitter Location Detection competition \citep{molla2014overview} was used to evaluate the approach for location reference recognition.
\citet{suat2022extraction} utilized a retrained spaCy NER to  detect and analyze traffic accidents from Spanish tweets in a city of Colombia.

%. Specifically, a Support Vector Machine (SVM) classifier was first trained to recognize tweets related to traffic accidents. A deep learning-based NER model (named \textit{es\_core\_news\_lg}) in spaCy was then retrained to recognize \textit{Location} and \textit{Time} entities in the tweets. Detected location references were then geocoded by using \textit{batch geocode} \footnote{\url{https://github.com/GISforHealth/batch_geocode}} that combines resources available on Google Map, OpenStreetMap, and GeoNames. 

Recently, many deep learning-based NERs have also been proposed. For example, \citet{limsopatham2016bidirectional} proposed recognizing name entities from tweets by enabling BiLSTM to automatically learn orthographic features using both the character embedding and word embedding. \citet{akbik2018contextual} proposed Flair, an NLP tool that used contextual string embeddings for sequence labeling tasks, such as part-of-speech (POS) tagging and NER. \citet{qi2020stanza} proposed a deep learning-based NLP toolkit, named Stanza, which adopted a contextualized string representation-based tagger. Recently, the fully-connected self-attention architecture (a.k.a. Transformer) attracts a lot of attention due to its parallelism and advantage in modeling long-range contexts. %Therefore, some studies adapted Transformer in NER tasks \citep{yan2019tener,ushio2021t}. %For instance,  proposed a transformer-based NER tool, named TENER. It adapted Transformer encoder to model the character-level features and word-level features. TENER was evaluated on CoNLL 2003 and OntoNoted 5.0 datasets. Aforementioned approaches utilized a fully supervised manner to train the model such that the performance of the model depends highly on the amount of training data, which is however difficult to annotate at large scales.
For instance, \citet{ushio2021t} presented a python library for NER model fine-tuning, named T-NER. It facilities the training and testing of a Transformer-based NER model. Nine public NER datasets from different domains are compiled as part of the T-NER library, such as CoNLL 2003, OntoNoted 5.0, and WNUT 2017 datasets.

\textbf{Learning-based PNE}: Apart from using or retraining existing NER models, many studies also trained their own models for location reference recognition by using  machine learning  \citep{sobhana2010conditional, nissim2004recognising, sagcan2015toponym} and deep learning models \citep{martinez2020reglas, chen2020location, kumar2019location, xu2019dlocrl, aldana2020adaptive, ushio2021t,cadorel2021geospatial,ma2022chinese}. For instance, \citet{nissim2004recognising} trained the Curran and Clark (C\&C) maximum entropy tagger \citep{curran2003language} for recognizing location references from Scottish historical documents, using the built-in C\&C features, including morphological and
orthographical features, information about the word itself, POS tags, named entity tag history (with a window size of 2), and contextual features. The model was evaluated on 648 Scottish historical documents, containing 10,868 sentences and 5682 places. % An F1 score of 0.94 is achieved by using 10-fold cross-validation.  
%\citet{sobhana2010conditional} trained a CRF model to recognize 17 classes of Named Entities (NEs), such as City, Island, River, Time, and Person, from Geological texts. Used features include prefix and suffix of length up to three of the current word, POS tags, digit features, information about the surrounding words and their NE tags. The approach was evaluated on a corpus of 200 scientific reports and articles on the geology of the Indian subcontinent by using 10-fold cross-validation. 
 % Recently, more and more studies  are leveraging deep learning to extract place names while avoiding feature engineering . % For instance, \citet{limsopatham2016bidirectional} proposed an approach for recognizing name entities from tweets by enabling bidirectional long short-term memory (BiLSTM) to automatically learn orthographic features using both the character embedding and word embedding. The model was trained on 2,349 tweets and tested on 3,850 tweets. % and achieved an F1-score of 0.66 on 11 entity types. The model achieved an F1-score of 0.73 on the location class. 
       %  For instance,
\citet{kumar2019location} implemented a multi-channel convolutional neural network (CNN) architecture to extract location references from tweets. The model was evaluated on 5107  earthquake-related tweets with 6690 place names by using 10-fold cross-validation. \citet{xu2019dlocrl} proposed DLocRL, a deep learning pipeline for fine-grained location recognition and linking in tweets. Specifically, they first used BiLSTM-CRF to train a  Point of Interest (POI) recognizer. Then, given an input pair ⟨POI, Profile⟩, a linking module was trained to judge whether the location profile corresponds to the POI. The profile is an entry in a POI dictionary. The approach was evaluated on the Singaporean national Twitter dataset that was first used in \cite{li2014fine}, containing 3611 tweets and 1542 POIs. % 2,500, 211 and 900 tweets were used for training, validation, and testing, respectively. %An F1 score of 0.83 is achieved in  location reference recognition. 
\citet{cadorel2021geospatial} proposed to extract a property's location and neighborhood from French housing advertisements by recognizing place names and retrieving relationships between them. Specifically, a BiLSTM-CRF network with a concatenation of several text representations (CamenBERT \citep{martin2019camembert}, Flair, and Word2Vec \citep{mikolov2013efficient}) was used to extract place names. % 1851 tweets were used for training and 930 tweets for testing. An F1 score of 0.81 was achieved.  
   % Using 10-fold cross-validation, the model achieved an F1-score of 0.96. 
    %Similarly, \citet{chen2020location} utilized a Bidirectional Long Short-Term Memory (BiLSTM) Neural Network to identify location information, especially aiming to recognize rarely known local places in social media, which are only familiar to locals. The model was trained on 4,215 tweets and tested on 100 tweets. The dataset corresponds to two consecutive typhoons, Hato and Pakhar in 2017 in the Southern China area.
    
To mitigate the effort for manually annotating a large training dataset, semi-supervised approaches have been developed. %, such as automatically generating training examples, or fine-tuning pretrained transformer models \citep{}. 
For instance, \citet{wang2020neurotpr} proposed to generate labeled training data from Wikipedia articles to train a BiLSTM model called NeuroTPR. Their model contains several layers to account for the linguistic irregularities in Twitter texts, such as the use of character embeddings to capture the morphological features of words, and contextual embeddings to capture the semantics of tokens in tweets. The approach was evaluated on 1000 tweets related to the 2017 Hurricane in Texas and Louisiana. \citet{qiuchinesetr}  proposed ChineseTR, a weakly supervised Chinese toponym recognizer. It first generated training examples based on word collections and associated word frequencies from various texts. Based on the training examples, a BiLSTM-CRF network built on BERT word embeddings was explored to train a toponym recognizer. The approach was evaluated on three Chinese NLP datasets (i.e., WeiboNER, Boson, and MSRA) \footnote{\url{https://github.com/InsaneLife/ChineseNLPCorpus}}. 
%The results show that the proposed ChineseTR achieves a 0.76 F1 score in a corpus with a 0.718 out-­of-­vocabulary rate and a 0.903 in-­vocabulary rate. All comparative ex- periments demonstrate that ChineseTR is an effective and scalable architecture that recognizes toponyms. 
 \citet{khanal2021multi} used a multi-task learning setting to augment the learning of fine-grained location identification. The  three tasks related to crisis events are key-phrase identification, eyewitness-account classification, and humanitarian category classification. The learning is conducted on one of the three popular transformer-based models: BERT \citep{devlin2018bert}, Albert \citep{lan2019albert}, and RoBERTa \citep{liu2019roberta}. Several public datasets for the training of the three tasks were utilized in the multi-task learning. The proposed approach was evaluated on two disaster-related twitter datasets that were used in \citet{middleton2018location}, which contain 1907 and 1762 tweets, respectively.
 
  Given abundant annotated data, statistical learning-based approaches can automatically recognize location references according to the contextual cues and the intrinsic features of location references without requiring additional expert knowledge and gazetteers. However, a large number of labeled training sentences are often not available, making it difficult to use these approaches in  many situations \citep{guerini2018toward}. Furthermore, deep learning based models normally take much more time to recognize place names from texts than rule and gazetteer matching-based approaches.

 %1,000 tweets from each are used as the training set and 500 tweets are used as the test set. The highest F1 score of 0.81 was achieved by the Albert-based approach.

   %To mitigate the challenge, \citet{guerini2018toward} proposed a domain portable zero-shot learning approach for entity recognition, which does not assume any annotated sentences at training time. More specifically, they trained a 3-layer BiLSTM model based only on available gazetteers and synthesized examples. It was then applied to recognize new entities in user utterances. Through multiple experiments in two languages (i.e., English and Italian) and three different domains (i.e., furniture, food, clothing), the proposed approach outperformed several competitive baselines, with minimal requirements of linguistic features. This work inspired the idea of our study.

 \subsubsection{Hybrid approaches}
 \label{Ha}

Every single technique has its own drawbacks. Thus, researchers have proposed fusing different techniques to achieve the best of all \citep{bontcheva2013twitie,li2014fine,malmasi2015location,weissenbacher2015knowledge, dutt2018savitr, hoang2018location, yenkar2021gazetteer}. %Table \ref{hybrid} summarizes hybrid approaches for location reference recognition, which 
Hybrid approaches can be further divided into four types based on the way they combine the previous three approaches: %according to the combination ways of the three techniques
{fusing rule and gazetteer}, {fusing rule and statistical learning}, {fusing gazetteer and statistical learning}, and {fusing rule, gazetteer, and statistical learning}.

\textbf{Fusing rule and gazetteer}: Many studies \citep{pouliquen2004geographical,weissenbacher2015knowledge,malmasi2015location,martinez2020reglas, wang2015spatiotemporal, moncla2014automatic, moncla2014geocoding} fused rules and gazetteers to overcome the shortcomings of each other. Manually defined rules are fragile and the detected location references can be thus further verified by gazetteers. Inversely, rules can be used to mitigate the two challenges faced by gazetteer matching, and can help remove the ambiguities of the location references detected by gazetteer matching and by recognizing those references that are not included in gazetteers. For instance, \citet{pouliquen2004geographical} proposed identifying cities and countries from newspapers in multiple languages. Location references are recognized by matching texts' n-grams written in upper case with a multi-language gazetteer, named Global Discovery gazetteer. %\footnote{From Europa Technologies Ltd, http://europa-tech.com/}. 
The matches are then filtered by stop words and person names. To recognize the morphological variants of places, regular expressions are used to list all possible suffixes and suffix combinations of location references. By doing so, some unseen places in gazetteers can be recognized, such as  \textit{`Lontoolaisen'}, because it consists of \textit{`Lontoo'} that is in the gazetteer and the suffix \textit{`laisen'}. To remove geo/geo ambiguities, several heuristics are utilized, such as discarding small places in the gazetteer, leveraging the importance of places, and determining the country of an article. The approach was evaluated on 28 texts with 1650 places in 8 languages (e.g., English, Spanish, and Russian). To help understand the origins, mutations, and the geospatial transmission patterns of viruses, such as influenza, rabies, and Ebola, \citet{weissenbacher2015knowledge} presented a geoparsing system for research articles related to phylogeography. GeoNames was first searched to detect location references in articles, and then a black-list (e.g., \textit{`How'}, \textit{`Although'}, \textit{`Gene'}, and \textit{`Body'}) and a set of rules were created to remove noisy entities found in GeoNames. %Specifically, linguistic rules were used to filter out acronyms, person names,   organization names, and adjectival uses of place names. The geo/geo ambiguity is removed by using the population heuristic, distance heuristic, and metadata heuristic (e.g., the \textit{country} field in a GenBank record linked to an article). The approach was tested on a corpus of 60 articles from PubMed Central, containing 1881 places. %An F1 score of 0.72 was achieved in place name extraction. 
\citet{malmasi2015location} proposed detecting location references in tweets. A POS rule-based tree splitting method is first used to extract noun phrases, and the n-grams of the noun phrases are then matched with the entries of GeoNames. %Furthermore, a list of directional and distance markers is compiled and REs are adopted to recognize complex location descriptions, such as \textit{`24 km North of Beijing'}. %An F1 score of 0.79 is achieved on 1,000 teeets. 
\citet{dutt2018savitr} presented SAVITR, a system that geoparses and visualizes tweets during emergencies. They used a POS tagger to find proper nouns, and then used REs to mitigate the ambiguity of proper nouns with the prefix and suffix words (e.g., \textit{`road'}, \textit{`south'}, and \textit{`city'}) of place names. Last, the phrases extracted by the above methods are verified and geocoded by using a gazetteer (i.e., GeoNames or OSM) in India. %An F1-score of 0.79 is achieved on 101 tweets when using GeoNames as the gazetteer.
    %Different from above studies that manually created rules, % that might not generalize well, 
\citet{martinez2020reglas} %derived rules from real examples. Specifically, the authors 
presented LORE, a knowledge-based model that captures location references from English and Spanish tweets. First, bi-grams and uni-grams in the tweets are matched with entries in the GeoNames gazetteer and then filtered by heuristics. %For instance, the first token of a bigram is not a noun and the unigram is in a stop word list. 
Second, linguistic patterns involving  location-indicative words (e.g., \textit{`city'} and \textit{`street'}), location markers  (e.g., \textit{`north'} and \textit{`10km'}), and POS tags are derived to recognize location expressions, such as  \textit{`25 miles NW of London City'}. They derived the linguistic patterns from 500 English tweets and 100 Spanish tweets, and then used 900 English tweets and 500 Spanish tweets to test LORE.

\textbf{Fusing rule and statistical learning}: Statistical learning models might not generalize well due to limited training samples, and manually defined rules can be added to boost the performance of the trained models, e.g., by correcting evident errors. %, such as removing unexpected detections and by including missing detections. 
 For instance, \citet{acheson2021extracting} proposed geoparsing scientific articles in the form of PDF by first recognizing candidate location references using Stanford NER and then filter the candidates using rules, such as to include candidates with \textit{`University'} or \textit{`Institute'} and reject candidates with \textit{`Inc'} and \textit{`GmbH'}. These rules are derived based on observations from the training sets. Google Geocoding API was then used to obtain the spatial representation of the detected location references. The approach was evaluated on two article corpora in the domain of Orchards and Cancer, containing 150 and 200 articles, respectively. %50 articles from each corpus were selected as test set and the remaining as training set. %In the step of location reference recognition, an F1 score of 0.75 and 0.8 were achieved on the Orchards and Cancer corpus, respectively.
\citet{das2019exploring} proposed detecting traffic events (e.g., traffic accidents and congestion) in India by using tweets. First, tweets are classified as traffic relevant
or not relevant using a supervised model. A hybrid method is then used to recognize location references in traffic-relevant tweets. Finally, the location references are geocoded by using OSM Nomination API \footnote{\url{https://github.com/osm-search/Nominatim}}. Specifically, the approach combines the detected location references by Stanford NER, retrained OpenNLP, and a rule-based system involving spatial indicators (e.g., \textit{`in'}, \textit{`at'}, and \textit{`near'}), POS tags, and 85 words of place categories (e.g., \textit{`hospital'}, \textit{`road'}, and \textit{`clinic'}).

 \textbf{Fusing gazetteer and statistical learning}: In this type of hybrid approaches, gazetteers are used in two main ways: (1) to combine the detection result of statistical learning models with gazetteer matching; (2) to use gazetteer matching result (e.g., if an n-gram is in gazetteers or not) as input features for statistical learning models. Examples of the first way are  \citep{freire2011metadata,habib2013hybrid,li2014fine,hoang2018location}. 
 For instance, to improve users' experience in searching their needed resources from digital libraries, \citet{freire2011metadata} proposed geoparsing descriptive metadata records associated with digital resources. Initial location references are recognized by matching tokens of records with candidate entries in GeoNames. A Random Forest classifier was then trained  to disambiguate and link the initial location references to the final entry. % in GeoNames. Some of the features that are used in the classifier are the case of the place name (e.g., uppercase or lowercase), if the place name was also found in a compiled collection of person names, the sum of the geographical distance to the other places found in the same record, and the population and type (e.g., Continent, Country, and City) of the places. 10-cross validation was performed on 752 records from Europeana, containing 2823 places.  
  \citet{li2014fine} proposed recognizing POIs in tweets. Candidate POIs in tweets were first extracted by matching with a POI inventory, which was constructed from check-in data in Foursquare. A trained time-aware POI tagger based on CRF was then utilized to remove the ambiguity of the candidates based on the context cues in the text. %The approach was evaluated on 4000 tweets using five-fold cross-validation. 
 \citet{hoang2018location} combined the detection results of multiple publicly available tools, such as Ritter's tool \citep{ritter2011named}, Gate NLP framework \citep{bontcheva2013twitie}, and Stanford NER, and then filtered the results using DBPedia. Different configurations of the NER tools and DBPedia were tested on the Ritter's dataset \citep{ritter2011named} and MSM2013 dataset \citep{cano2013making}. 
Examples of the second way include  \citep{inkpen2017location, weissenbacher2017extracting, magge2018deep, peterson2021automated,nLORE}. For instance, \citet{inkpen2017location} trained three CRF models for recognizing city, province/state, and country mentions based on manually defined features, including gazetteer features (if a phrase is in GeoNames or not). The models are intended to not only detect location references in tweets but also to categorize them into three types. The models were evaluated by using 10-fold cross-validation on 6000 tweets, containing 1270 countries, 772 states or provinces, and 2327 cities. %The results showed the F1 score for countries, states or provinces, and cities is 0.92, 0.85, and 0.83, respectively.
To support viral phylogeographic studies, \citet{weissenbacher2017extracting} proposed recognizing location references in research articles pertaining to virus-related GenBank records by using a CRF model. Lexical (i.e., POS tags), semantic, and gazetteer features. The proposed approach was evaluated on the same dataset as \citet{weissenbacher2015knowledge}.     \citet{nLORE} proposed nLORE, a BiLSTM-CRF architecture for location reference recognition, exploiting both linguistic and gazetteer features from LORE \citep{martinez2020knowledge}. %and semantic information contained in word embeddings. 
The model was trained on 7000 tweets and then tested on 1063 tweets. %, which is an annotated corpus of 60 articles with 1,881 places from PubMed Central. %48 articles with 1,596 places were used as training set and 12 articles with 285 places were used as test set. %A F1 score of 0.81 is achieved.
 
\textbf{Fusing rule, gazetteer, and statistical learning:} Some studies combined all the three techniques for location reference recognition \citep{lieberman2011multifaceted,gelernter2013cross,magge2018deep,dutt2019utilizing,hu2020GazPNE,gazpne2}. For instance, %\citet{lieberman2011multifaceted} proposed a hybrid approach for  recognizing toponyms from streaming news. They combined a dictionary of entity (e.g., \textit{Location}, \textit{Person}, and \textit{Organization}) names and spatial cue words (e.g., \textit{`County of'}), POS tags, statistical learning based NER (i.e., Stanford NER), rule-based toponym refactoring, and grammar filters involving noun adjuncts and active verbs (e.g., location does not perform actions). To prepare the gazetteer, continents, countries, states/provinces, cities, and counties as well as their abbreviations are retrieved from GeoNames. Rule-based toponym refactoring is used to deal with the toponym variants, such as to expand \textit{`County Kildare'} to \textit{`Kildare County'} and \textit{`County of Kildare'}, and to expand \textit{`Ft. Meade'} to \textit{`Fort Meade'}. The method was evaluated on two news article datasets: LGL \citep{lieberman2010geotagging} and Clust which is annotated by the authors, including 1,701 articles and 16,329 toponyms in total. %A F1 score of 0.75 is achieved by the proposed method.
\citet{gelernter2013cross} proposed a cross-lingual location reference recognition for tweets, combining the results of a named
location parser by using gazetteer matching, a rule-based building parser, a rule-based street
parser, and a trained CRF-based named entity parser. The rules of the street and building parsers are created based on POS tags and indicator words, such as adjective plus noun and street and building indicators (e.g., \textit{`street'} and \textit{`highway'} in English and \textit{`calle'} and \textit{`carreterra'} in Spanish). 4488 Spanish crisis-related tweets with 3182 tweets as training set and the rest as test set are used to evaluate the Spanish extractor. The Spanish dataset is then translated into English with the Google translator and used to evaluate the English extractor. %The results show that an F1 of 0.8 and 0.86 is achieved by the Spanish and English extractors, respectively.
 \citet{magge2018deep} used a deep feedforward neural
network to determine whether a given phrase in biomedical articles is a toponym or not. Rules are used to generate 8 million training samples from unannotated datasets. The generated samples are then used together with manually annotated training samples to train a model. %, which are then evaluated on manually annotated test datasets. The manually annotated training and test sets are the same as \citet{weissenbacher2017extracting}. 
 The phrase's context (represented by word embeddings), properties of the phrase (e.g., if it is in GeoNames), properties of the document (i.e., {abstract}, {introduction}, {body}, or {table}) are concatenated into the input vector of the deep learning model. %The results showed that the F1 score increased from 0.91 to 0.93 when synthesized training samples are used.
 \citet{dutt2019utilizing} proposed understanding five important aspects of need-tweets and availability-tweets during disasters, including what resource (e.g., water, food, shelter, and medicines) and what quantity is needed/available, the geographical location of the need/availability, and who needs/is providing the resource. With regard to geoparsing, the authors improved their previously proposed system, Savitr \citep{dutt2018savitr}, by combining the location references detected by spaCy and a rule system and then % that uses POS tags, spatial propositions, and words of place categories. % and India and non-place names that are synthesized by rules. The neural classifier was then utilized to the valid n-grams . 
 %The detected place names are then 
 filtering the location references %and geocoded 
 through a gazetteer. %Based on the proposed geoparsing approach, a utility-driven model for optimized resource allocation in a post-disaster scenario based on real-time tweets was proposed \citep{basu2022utilizing}.
 More recently, we proposed two place name extractors for tweets. The first extractor is named GazPNE \citep{hu2020GazPNE}, which is a neural classifier  first trained based on place names in OpenStreetMap in the region of the US and India and non-place names that are synthesized by rules. %The classifier was then applied to the valid n-grams of a tweet, which satisfy POS rules. %GazPNE was evaluated on three flood-related twitter datasets that were initially utilized in \citep{al2017location}. 
 Because GazPNE still suffers from ambiguity issues due to its limited use of context information, we developed a second and more robust approach, GazPNE2 \citep{gazpne2}. It utilizes two pretrained transformer models, BERT and BERTweet \citep{nguyen2020bertweet} to disambiguate the detected location references and achieved an improved F1 score of 0.8 on 19 public twitter datasets. % and an F1 score of 0.8 was achieved.
 
% Hybrid approaches attract a lot of attentions among researchers. Many novel hybrid approaches have been proposed to overcome the shortcomings of each single approach. Some of the representative hybrid approaches will be selected and evaluated on a large benchmark data.

\subsection{Comparative studies}
In addition to individual studies that focused on developing new methods, researchers also conducted experiments to compare existing methods based on the same datasets. \citet{liu2014automatic} created a medium-scale corpus of locative expressions from multiple social media sources which include the TellUsWhere corpus \citep{winter2011location}, two sets of micro-blog posts from Twitter, comments from YouTube, forums, blog posts from tier one of the ICWSM-2011 Spinn3r dataset \footnote{\url{https://www.icwsm.org/2011/data.php}}, Wikipedia, and documents from the British National Corpus \citep{burnard1995users}. %Each corpus contains 500 annotated sentences. 
They then compared the performance of a couple of location reference recognition models over these seven corpora, which include Locative Expression Recogniser (LER) \citep{liu2013automatic}, retrained Stanford NER, pretrained Stanford NER, GeoLocator \citep{gelernter2013algorithm}, UnLockText, and Twitter NLP. %UnLockText is developed by the Language Technology group of the University of Edinburgh. The link to the tool is unavailable anymore. 
The results show the pretrained Stanford NER achieves the best overall performance. 
\citet{gritta2018s} evaluated the performance of five geoparsers (GeoTxt, Edinburgh Geoparser \citep{grover2010use}, Yahoo! PlaceSpotter, CLAVIN, and Topocluster \citep{delozier2015gazetteer}) on two datasets, Local-Global Lexicon (LGL) \citep{lieberman2010geotagging}, and WikToR that was programmatically created by the author. For location reference recognition, GeoTxt uses Stanford NER, Edinburgh Geoparser uses LT-TTT2, TopoCluster uses Stanford NER, and CLAVIN uses Apache OpenNLP. The evaluation results showed that Stanford NER performs the best in location reference recognition, and Edinburgh Geoparser and CLAVIN perform the best in geocoding.
\citet{wang2019enhancing}  developed an extensible and unified platform for evaluating geoparsers, named EUPEG, which enables direct comparison of nine geoparsers on eight public corpora, %containing 13,349 locations in total. 
which are LGL, GeoVirus \citep{gritta2018melbourne}, TR-News \citep{kamalloo2018coherent}, GeoWebNews \citep{gritta2018pragmatic}, WikToR \citep{gritta2018s}, GeoCorpora \citep{wallgrun2018geocorpora}, Hu2014 \citep{hu2014improving}, and Ju2016 \citep{ju2016things}.  %Note that, except GeoCorpora, which is a twitter dataset, the others are formal texts. 
The compared geoparsers include GeoTxt, Edinburgh Geoparser, TopoCluster, CLAVIN , Yahoo! PlaceSpotter, CamCoder \citep{gritta2018melbourne} that uses spaCy NER for location reference recognition, DBpedia Spotlight \citep{mendes2011dbpedia}, and two systems that use Stanford NER and spaCy NER for location reference recognition, respectively, and population-based heuristics for disambiguation in geocoding step. %The evaluation result of the location reference recognition step shows that Yahoo! PlaceSpotter achieves the best  F1 score of 0.73.
\citet{won2018ensemble} evaluated the performance of five NERs and voting systems that combine the NERs in extracting place names from two collections of historical correspondence, named Mary Hamilton Papers and the Samuel Hartlib collection. The NERs include NER-Tagger \citep{lample2016neural}, Stanford NER, spaCy, Edinburgh Geoparser, and Polyglot-NER \citep{al2015polyglot}. The results showed that although the individual performance of each NER system was corpus dependent, the ensemble combination can achieve consistent measures of precision and recall, outperforming the individual NER systems. At the International Workshop on Semantic Evaluation 2019 \footnote{\url{https://alt.qcri.org/semeval2019/}}, a task for toponym resolution in scientific articles was launched. The evaluation results were presented in \citep{weissenbacher2019semeval}. Several systems were evaluated on a corpus of 150 full PubMed articles as 105 articles for training and 45 articles for testing, containing in total 8360 toponyms. In the subtask of toponym recognition, all systems except one adopted Deep Recurrent Neural Networks. %Their architectures varied with respect to the integration of character embedding layers, mechanisms of attention, integration of external features (e.g., POS tags or other Named Entities) or the choice of a general or in domain corpus for pre-training their word and sentence embeddings. 
The highest F1 score was achieved by the system proposed by a team from Alibaba Group, which adopted an architecture of BiLSTM-CRF that was trained on OntoNote5.0, CoNLL13, and weakly labeled training corpora. All systems relied on handcrafted features for toponym resolution, including the lexical context of the toponyms and their importance (e.g., population), and a gradient boosting algorithm performs the best.

There are two major differences between this study and the aforementioned comparative studies. First, these existing comparative studies focused on the entire workflow of geoparsing, while we focus on a narrower topic, i.e., location reference recognition, and provide a deeper review and comparison of methods on this topic. Second, our comparative experiments (presented in the following section) are more comprehensive than existing studies. We used more datasets (26 datasets, containing 39,736 places across the world) and compared \toolss\thinspace  different approaches. In the following, we present the results from the comparative experiments.

% including both classic and the newest NER tools and location-specific tools.

\section{Comparison of existing approaches}
\label{method}
	  \subsection{Methods}
	  
	  \begin{table}[]
\footnotesize
\centering
\caption{Main features of tools evaluated in this study} \label{tools}
\begin{tabular}{|c|c|c|c|c|c|}
\hline
Tool and Version  &   Category    & \begin{tabular}[c]{@{}c@{}}Recognized \\ Entity Type\end{tabular} & Target Data  & \begin{tabular}[c]{@{}c@{}}Development\\ Language\end{tabular} & \begin{tabular}[c]{@{}c@{}}Publishing\\ Time\end{tabular} \\ \hline
Stanford NER 4.3.1  &  statistical learning  & 4 classes                                                  & formal texts & Java                                                           & 2021-11                                                   \\ \hline
spaCy 3.2.1    &  statistical learning          & 18 classes                                                  & formal texts & Python                                                         & 2021-12                                                   \\ \hline
Stanza 1.2   &  statistical learning            & 18 classes                                 & formal texts & Python                                                         & 2021-01                                                   \\ \hline
OpenNLP 1.9.4   &  statistical learning         & 4 classes                                                    & formal texts & Java                                                           & 2021-11                                                   \\ \hline
DBpedia Spotlight  &  statistical learning      & *                                                    & formal texts & Python                                                         & 2021-02                                                   \\ \hline
NER-Tagger      &  statistical learning         & 4 classes                                                   & tweets       & Python                                                         & 2016-04                                                   \\ \hline
Polyglot 16.07.04  &  statistical learning      & 3 classes                                                 & formal texts & Python                                                         & 2016-07                                                   \\ \hline
NeuroNER         &  statistical learning        & 4 classes                                                      & formal texts & Python                                                         & 2017-05                                                   \\ \hline
CogComp 4.0     &  statistical learning         & 4 classes                                                      & formal texts & Java                                                         & 2018-08                                                   \\ \hline
OSU TwitterNLP  &  hybrid        & 10 classes                                                   & tweets       & Java                                                           & 2011-07                                                   \\ \hline
TwitIE-Gate 9.0.1   &  hybrid     & 4 classes                                                     & tweets       & Java                                                           & 2013-09                                                   \\ \hline
% TENER-CoNLL             & 4 classes                                                         & formal texts & Python                                                         & 2019-11                                                   \\ \hline
% TENER-Ontonotes         & 11 classes                                                        & formal texts & Python                                                         & 2019-11                                                   \\ \hline
TNER          &  statistical learning           & 28 classes                                       & formal texts & Python                                                         & 2021-04                                                   \\ \hline
Flair NER       &  statistical learning         & 4 classes                                                  & formal texts & Python                                                         & 2021-02                                                   \\ \hline
Flair NER (Ont)  & statistical learning    & 18 classes                                 & formal texts & Python                                                         & 2021-03                                                   \\ \hline
BERT-base-NER    &  statistical learning        & 4 classes                                                 & formal texts & Python                                                         & 2020-04                                                   \\ \hline
CLIFF 2.6.1      &  statistical learning      & LOC                                                    & formal texts & Python                                                         & 2020-04                                                   \\ \hline
Edinburgh 1.2 &  hybrid  & LOC                                                           & formal texts & c                                                         & 2021-07                                                   \\ \hline

GazPNE2      &  hybrid            & LOC                                                       & tweets       & Python                                                         & 2022-02                                                   \\ \hline

LORE        &  hybrid             & LOC                                                       & tweets       & c++                                                            & 2020-12                                                   \\ \hline
nLORE      &  hybrid              & LOC                                                        & tweets       & c++                                                            & 2021-06                                                   \\ \hline
SPENS     &  hybrid              & LOC                                                           & formal texts & N/A                                                            & 2018-03                                                   \\ \hline
RSD       &  hybrid               & LOC                                                          & tweets       & N/A                                                            & 2018-03                                                   \\ \hline
RGD       &  hybrid               & LOC                                                      & tweets       & N/A                                                            & 2018-03                                                   \\ \hline
RS        &  hybrid               & LOC                                                          & tweets       & N/A                                                            & 2018-03                                                   \\ \hline

BaseSemEval12  &  hybrid          & LOC                                                        & formal texts & Python                                                         & 2018-09                                                   \\ \hline
NeuroTPR       &  statistical learning           & LOC                                                           & tweets       & Python                                                         & 2020-10                                                   \\ \hline
Geoparserpy 2.1.4   &  gazetteer matching     & LOC                                                        & tweets       & Python                                                         & 2020-06                                                   \\ \hline
% Georeferencing          & LOC                                                               & tweets       & Java                                                           & 2020-11                                                   \\ \hline
% CNN-based               & LOC                                                               & tweets       & N/A                                                            & 2018-12                                                   \\ \hline

\end{tabular}
\end{table}

To inform future methodological developments for location reference recognition and help guide the selection of proper approaches based on application needs, we examine numerous characteristics of existing approaches for location reference recognition. We use or implement the \toolss\thinspace most widely used approaches including both general NERs and location-specific approaches. Note that, we do not include several approaches, such as LNEx \citep{al2017location}, GazPNE \citep{hu2020GazPNE}, and SAVITR \citep{dutt2018savitr}, since they can only be applied to a local region rather than the entire globe, while the place names of our test datasets are across the globe. Table \ref{tools} summarizes the features of the compared approaches. The number after the name of a tool represents its version. The second column denotes the category of the approach with regard to its underlying principle. NERs can not only recognize \textit{Location} but also other entity types, as shown in the third column. 3 classes, 4 classes, 10 classes, and 18 classes denote \{LOC, PER, ORG\}, \{LOC, PER, ORG, MISC\}, \{PERSON, GEO-LOCATION, COMPANY, PRODUCT, FACILITY, TV-SHOW, MOVIE, SPORTSTEAM, BAND, OTHER\}, and \{LOC, PERSON, ORG, FAC, GPE, CARDINAL, DATE, EVENT, LANGUAGE, LAW, MONEY, NORP, ORDINAL, PERCENT, PRODUCT, QUANTITY, TIME, WORK\_OF\_ART\}, respectively. 28 classes includes the entities of the 18 classes and the entities of \{CELL TYPE, CELL LINE, CHEMICAL, CORPORATION, DISEASE, DNA, GROUP,  PROTEIN, RNA, OTHER\}. Note that, DBpedia Spotlight can recognize quite a few detailed classes, such as Place, People, Event, and Color. %The fourth column lists the types recognized as locations in the evaluation. 
The fourth column refers to the type of datasets on which it was developed. The fifth column refers to the development language of the approach. \textit{N/A} means the code of the approach is unavailable and we reimplement the approach by ourselves. The last column refers to the time that a tool of a certain version was proposed or updated.  %The fifth column denotes if an approach was designed for dealing with the location mentions in complex hashtags, such as \textit{`Marysville'} in \textit{`\#MarysvilleShooting'}. The last column denotes the year that a tool of a certain version was proposed or updated. 
%A list of the compared tools or approaches are summarized as below. %Some tools have multiple versions. 
%The number in the parentheses after the tool name indicates its version. By default, we use the latest version of a tool.
 
    %   \textbf{Google NLP} \footnote{\url{https://cloud.google.com/natural-language/}}: It is a general-purpose NER tool deployed at the Google Cloud. We considered its outputted entities mentioned as LOCATION, ADDRESS as locations.
       
\begin{itemize}
    \item \textbf{Stanford NER (4.3.1) } \citep{finkel2005incorporating} : It is a Java implementation \footnote{\url{https://nlp.stanford.edu/software/CRF-NER.html}} of a NER based on CRF, which was developed and maintained by the Stanford Natural Language Processing Group. We kept the entities of LOC (location) detected by Stanford NER as locations.

    \item   \textbf{spaCy (3.2.1)}: It is a general NLP tool. %\footnote{\url{https://spacy.io/}} and offers a fast statistical NER approach. 
      %The latest version  (3.0) of spaCy is based on transformer. 
      We used its retrained model (\textit{en\_core\_web\_lg}) and  kept the entities of LOC, FAC (facility), and GPE (geopolitical entity) detected by spaCy as locations.
       
            \item   \textbf{Stanza (1.2)} \citep{qi2020stanza}: It is a general NLP toolkit \footnote{\url{https://stanfordnlp.github.io/stanza/}} and includes an NER tool, which was built on BiLSTM and CRF. We kept the entities of LOC, FAC, and GPE as locations.
              
   \item    \textbf{OpenNLP (1.9.4)} \citep{mendes2011dbpedia}: The Apache OpenNLP library %\footnote{\url{https://opennlp.apache.org/}}
      is an open sourced and  machine learning based toolkit for the processing of natural language text. We kept the entities of Location detected by OpenNLP as locations.

  \item     \textbf{DBpedia Spotlight} \citep{mendes2011dbpedia}: It is for recognizing and linking entities based on a knowledge base--DBpedia. We treated the place mentions detected by this tool \footnote{\url{https://www.dbpedia-spotlight.org/}} as locations in the evaluation.

 \item      \textbf{NER-Tagger} \citep{lample2016neural}: It is a NER tool  targeted for tweets. It was built on BiLSTM and CRF. We used the pretrained model and implementation \footnote{\url{https://github.com/glample/tagger}} to detect locations that were tagged with B-LOC and I-LOC. %If an entity tagged with I-Loc is follows another entity tagged with B-Loc or I-Loc, these two entities are combined as one location.

 \item    \textbf{Polyglot (16.07.04)} \citep{al2015polyglot}: It is a natural language pipeline \footnote{\url{https://polyglot.readthedocs.io/en/latest/index.html}} and includes a multi-language NER tool. The entities tagged with I-LOC by this tool were regarded as locations.
 
\item         \textbf{NeuroNER} \citep{dernoncourt2017neuroner}: It is a BiLSTM-CRF based NER system developed by MIT. We used the pretrained model and implementation \footnote{\url{https://github.com/Franck-Dernoncourt/NeuroNER}} to tag entities, and the locations were those with the tags of LOC, FAC, and GPE.

  \item   \textbf{CogComp (4.0)} \citep{ratinov2009design}:
    It is a NER tagger \footnote{\url{https://github.com/CogComp/cogcomp-nlp/tree/master/ner}}, which was developed by the University of Illinois. The entities tagged with LOC were taken as the locations identified by this tool.

   \item        \textbf{OSU Twitter NLP} \citep{ritter2011named}: It is a twitter-specific NER tool \footnote{\url{https://github.com/aritter/twitter_nlp}} that particularly targets twitter texts. The entities tagged with GEO-LOCATION and FACILITY by this tool were treated as locations.
           
     \item              \textbf{TwitIE-Gate (9.0.1)} \citep{bontcheva2013twitie}: It is a twitter-specific NER tool \footnote{\url{https://gate.ac.uk/wiki/twitie.html}},  providing an executable pipeline on an open-source software toolkit GATE \footnote{\url{https://gate.ac.uk/}} (General Architecture for Text Engineering). The entities tagged with {Location} by this tool were treated as locations.

%     \textbf{TENER-CoNLL} \citep{yan2019tener}: TENER \footnote{\url{https://github.com/fastnlp/TENER}} (Transformer Encoder for Named Entity Recognition) is a Transformer-based model which aims to tackle the NER task. We retrain a TENER model based on the CoNLL2003 dataset.
  
%   \textbf{TENER-OntoNotes} \citep{yan2019tener}:  We retrain a TENER model based OntoNotes-5.0 NER dataset.
    
\item     \textbf{TNER} \citep{ushio2021t}: It is an All-Round Python Library  \footnote{\url{https://github.com/asahi417/tner}} for Transformer-based Named Entity Recognition. Its recognized locations included those entities tagged with LOC, GPE, and FAC.

\item       \textbf{Flair NER} \citep{akbik2019flair}: Flair is an NLP framework designed to facilitate training and distribution of sequence labeling and text classification. Flair-NER is the standard 4-class NER model trained on CoNLL-03. We used their trained model  \footnote{\url{https://huggingface.co/flair/ner-english}} directly and identified locations through the LOC tag.

  \item     \textbf{Flair NER (Ontonotes)} \citep{schweter2020flert}: This is the large 18-class NER model trained  on Ontonotes that ships with Flair. It is named \textit{Flair NER (Ont)} for short in this review. We used their trained model  \footnote{\url{https://huggingface.co/flair/ner-english-ontonotes-large}} directly and included entities tagged with LOC, GPE, and FAC as locations. 
       
   \item    \textbf{BERT-base-NER}: It is a fine-tuned BERT model that is ready to use for Named Entity Recognition. We used their trained NER model \footnote{\url{https://huggingface.co/dslim/bert-base-NER}} directly. The locations were derived from the entities tagged with B-LOC and I-LOC. %For such a IOB (inside, outside, beginning) format, we regarded the combination of consecutive entities that start with B- format and ends with I- format as a single location. 

  \item     \textbf{GazPNE2}
         \citep{gazpne2}: It fuses global gazetteers and two pretrained transformer models. The latest version \footnote{\url{https://github.com/uhuohuy/GazPNE2}} utilized Stanza to accelerate GazPNE2 and detect hard examples for GazPNE2. 

 \item      \textbf{CLIFF (2.6.1)} \citep{d2014cliff}: It  integrates the results of Stanford NER  and a modified CLAVIN (Cartographic Location and Vicinity Indexer) geoparser. % \footnote{\url{https://github.com/Novetta/CLAVIN}}. 
      In the evaluation, we used its implementation \footnote{\url{https://cliff.mediacloud.org/}} and kept the detected place mentions as locations.
      
\item       \textbf{LORE} \cite{martinez2020knowledge}: It is  a rule-based location extractor for tweets. We used its implementation to extract locations. 

  \item     \textbf{nLORE} \cite{martinez2020reglas}: It is a deep learning model, an advanced version of \textbf{LORE}. We used the trained model provided by the author to extract locations.

 \item      \textbf{Edinburgh Geoparser (1.2)} \citep{grover2010use}: It is a geoparsing tool \footnote{\url{http://www.ltg.ed.ac.uk/software/geoparser/}}  developed by Edinburgh University, which combines rules and gazetteers to extract place names directly from text.
       
 \item      \textbf{BaseSemEval12} \citep{magge2018deep}: It is a baseline system for SemEval-2019 Task 12 (i.e. Toponym Resolution in Scientific Papers) that uses a 2-layer feedforward neural network \footnote{\url{https://github.com/amagge/semeval-ffnn-baseline}}. Its output place mentions were taken as the detected locations.

  \item     \textbf{NeuroTPR} \citep{wang2020neurotpr}: It is a neuro-net toponym recognition tool trained on recurrent neural networks. We  used their trained model  and implementation \footnote{ \url{https://github.com/geoai-lab/NeuroTPR}} to detect location mentions in texts.
              
 \item      \textbf{Geoparserpy (2.1.4)} \citep{middleton2018location}: It is a  representative gazetteer-based geoparser. We used the implementation of Geoparserpy and deployed the required  OpenStreetMap gazetteer to extract place names.
       
%   \item     \textbf{Georeferencing} \citep{das2019exploring}: It combines the result of Stanford NER, retrained OpenNLP, and a rule system to recognize place names. We use its implementation\footnote{\url{https://github.com/rddspatial/georeferencing}} to tag the evaluation dataset.

   \item    \textbf{SPENS} \citep{won2018ensemble}: This approach combines the result of five different systems in a voting mechanism, including Stanford NER, Polyglot NER, Edinburgh Geoparser, NER-Tagger, and spaCy. It is thus named SPENS for short. We reimplement the approach by using the code and API of the five modules.% The location tagging results from these five systems were aggregated first, after which they were voted through based on an agreement threshold among these five systems. In this review, the threshold was set as two, when the results achieved the highest evaluation metrics.

      % \textbf{CNN-based approach} \citep{kumar2019location}:  The CNN-based approach is re-implemented and re-trained. We use the parameters suggested by \citet{kumar2019location} and also changed them to obtain the best training result. 
    %   We divide the datasets into two groups, with one group containing datasets $a, c, e,g,i,k,m,o,q,s$ and the other group containing datasets $b,d,f,h,j,l,n,p,r$. One Group is used as training data and the other is used as the test data, and vice versa. This is to evaluate the transferability of the trained model. The training setting is reasonable since in a real situation it cannot be guaranteed that the new data overlaps with the training data.

\item      \textbf{Ritter+Stanford NER+DBpedia} \citep{hoang2018location}: It uses DBpedia to filter the merged detection by Ritter's tool (also named OSU Twitter NLP) and Stanford NER. We name this approach RSD for short and reimplement the approach by using the code and API of the three modules.
       
  \item     \textbf{Ritter+GATE+DBpedia}
      \citep{hoang2018location}: It uses DBpedia to filter the merged detection by Ritter's tool and GATE. We name this approach RGD for short and reimplement the approach by using the code and API of the three modules. 
       
 \item   \textbf{Ritter+Stanford NER}
         \citep{hoang2018location}: It merges detection by Ritter's tool and Stanford NER. We name this approach RS for short and reimplement the approach by using the code and API of the two modules. 
       
\end{itemize}

All methods were configured taking into account the corresponding research results to ensure to choose the optimal parameter settings. For example, we consider not only \textit{Location} and \textit{GPE} but also \textit{Facility} detected by Stanza as a location since this can achieve the best F1 score on the total datasets.

% 	  % ##########################################
	  
	  \subsection{Test data}	    
	  We collect in total 26 commonly used datasets and use them as test data. The datasets include 3 formal datasets (i.e., news) and 23 informal datasets (i.e., tweets), containing 39,736 place names in total, as shown in Table \ref{dataset}. %The datasets are from different sources, such as tweets, Wikipedia articles, web pages, and news. 
         They can be categorized into two groups based on the purpose of the datasets: Location Extraction (LE) and NER. The former only annotates \textit{Location} while the latter annotates not only \textit{Location}, but also the other types, such as \textit{Person}, \textit{Organization}, and \textit{Facility}. Note that, we do not use some available geoparsing datasets that were used to evaluate the geoparsing approaches in \citep{wang2019enhancing}, such as WikToR \citep{gritta2018s} since we found they miss quite a few toponyms. For example, in WikToR, a text or article corresponds to a WikiPedia page entitled with a toponym whose coordinates are specified. The text explains the toponym. Only this toponym is automatically annotated, ignoring the other toponyms in the text. The dataset can be used to evaluate toponym resolution approaches but not toponym recognition approaches. The description of the used datasets is as follows:
         \begin{itemize}
              
         \item \textbf{LaFlood2016, HouFlood2015, CheFlood2015  \footnote{The datasets can be obtained by filling out the Dataset Registration form \url{https://docs.google.com/forms/d/e/1FAIpQLScf6-DNwkgJXPS5e28Mj18hIW3Ap_Ym7Kna-SO7oSmiC72qGw/viewform}}}: They are three flood related datasets, which were created by \citet{al2017location}. The locations in the three datasets were annotated as one of the three types: \textit{inLOC}, \textit{outLOC}, and  \textit{ambLOC}, denoting the locations inside the area (e.g., \textit{`Houston'}) of interest, outside the area, and ambiguous locations (e.g., \textit{`my house'}), respectively. We only evaluate the tools on the \textit{inLOC} and \textit{outLOC} locations, ignoring the \textit{ambLOC} locations. \textit{Louisiana}, \textit{Houston}, \textit{Texas}, and \textit{Chennai}, as well as their abbreviations, such as \textit{`La'}, \textit{`Hou'}, and \textit{`Tx'} appear frequently in the datasets. Moreover, many location mentions are in hashtags, such as \textit{`\#\textbf{la}flood'}, \textit{`\#\textbf{tx}wx'}, and \textit{`\#\textbf{Chennai}Rain'}.

   \item \textbf{Harvey2017 \footnote{ \url{https://github.com/geoai-lab/NeuroTPR/tree/master/Data/TestData/HarveyTweet2017}}}: The dataset is related to 2017 Hurricane Harvey and was created by \citet{wang2020neurotpr}. The dataset contains many fine-grained locations, such as \textit{`398 Garden Oaks Blvd'} and \textit{`26206 longenbaugh rd'}. No places appear in hashtags since they have been removed from the dataset.  
     
    \item \textbf{ NzEq2013, NyHurcn2012 \footnote{\url{https://revealproject.eu/geoparse-benchmark-open-dataset/}}}: The two twitter datasets correspond to the New Zealand earthquake in 2013 and New York Hurricane in 2012, respectively. They were created by \citet{middleton2018location}. We found several missing  place names (e.g., \textit{`Christchurch'}) which, however, appear frequently in the two datasets. %For instance, in the text \textit{`Time-lapse visualisation of todays earthquakes in Christchurch and Canterbury, New Zealand.'}, \textit{`New Zealand'} is not annotated in dataset $f$. 
     To mitigate this issue, we manually create two missing place name lists (i.e. [(\textit{`new',`zealand'}), (\textit{`nz'}), (\textit{`uk'}), (\textit{`christchurch'}), (\textit{`chch'}), (\textit{`lyttleton'}), (\textit{`southland'}), (\textit{`wellington'}), (\textit{`south', `island'})] and [(\textit{`new',`york'}), (\textit{`nyc'}), (\textit{`new',`york',`city'}), (\textit{`ny'})] ) for the two datasets, respectively. We define that the detection of an entity which is not annotated in the dataset but in the corresponding missing list is a true positive. Moreover, sub place names exist in dataset NzEq2013. For example, in the text \textit{`Christchurch hospital is now back in operation'}, both \textit{`Christchurch hospital'} and \textit{`Christchurch'} were annotated as \textit{Location}. To tackle this issue, we removed sub place names from the dataset. %{\color{red} 100 such cases have been founded.} 
     
     \item \textbf{Martinez\_I, Martinez\_II, Martinez\_III:} The three twitter datasets correspond to multiple crises and emergency events (e.g., earthquake, flood, car accident, bombing, shooting, terrorist, and incident) that happened across the world. They were initially utilized in \citep{martinez2020knowledge,nLORE}. One of the features of the datasets is that many fine-grained locations, such as \textit{`13219 S penrose Ave'} and \textit{`Exit 34'} as well as complex location expressions, such as \textit{`50 miles SW of Liverpool}' and \textit{`25mins away from Northumbria Street'} were annotated.

      \item \textbf{GeoCorpora\footnote{\url{https://github.com/geovista/GeoCorpora}}:} It was created by \citet{wallgrun2018geocorpora}. In the dataset, location references in tweets were not only annotated but also linked to the toponyms of GeoNames. Therefore, it can be also used to evaluate the geocoding approaches. The dataset corresponds to multiple noteworthy events (e.g., earthquake, ebola, fire, flood, protest, and rebel) that happened across the world in 2014 and 2015. The majority of the annotated places are admin units, such as continent, country, state, and city. 

\item \textbf{CrisisBench-1000, HumAID-1000, COVID19-1000 \footnote{\url{https://github.com/uhuohuy/GazPNE2/tree/main/data/test_data}}}: The three datasets were created by ourselves. Specifically, we randomly selected 1000 tweets from CrisisBench \cite{alam2020crisisbench}, HumAID \cite{humaid2020}, and a COVID19 dataset \cite{lamsal2021design}, respectively, and then manually annotated place names, including admin units (e.g., country and village), traffic ways (e.g., street and highway), natural features (e.g., hill and river), and POIs (e.g., park and school).

     \item \textbf{BTC-A, BTC-B, BTC-E, BTC-F, BTC-G, BTC-H \footnote{\url{https://github.com/GateNLP/broad_twitter_corpus}}}: The Board Twitter Corpus (BTC) was created by \citet{derczynski2016broad}. The datasets were sampled across different regions, temporal
periods, and types of Twitter users. Apart from \textit{Location}, \textit{Organization} and \textit{Person} were also annotated. Several annotated place names are in mentions (e.g., \textit{`@HoustonFlood'}). However, they are normally ignored by existing location extractors. Thus, we remove the place name in the mentions from the six datasets. 
     
     \item \textbf{NEEL2016 \footnote{\url{http://microposts2016.seas.upenn.edu/challenge.html}}}: It is the gold dataset  of 2016 Named Entity rEcognition and Linking (NEEL) Challenge. The dataset includes event-annotated tweets covering multiple noteworthy events from 2011 to 2013, such as the death of Amy Winehouse, the London Riots, the Oslo bombing, and the Westgate Shopping Mall shootout. Entities of different types, such as \textit{Location}, \textit{Person}, \textit{Organization}, \textit{Event}, and \textit{Product} were not only annotated but also linked to entries in DBPedia. We used its training set, which contains 2135 tweets and 602 places.

    \item \textbf{Ritte’s dataset: \footnote{\url{https://github.com/aritter/twitter_nlp/blob/master/data/annotated/ner.txt}}} It was initially used by \citet{ritter2011named}. \textit{Location}, \textit{Facility}, \textit{Person}, and \textit{Organization} were annotated in the dataset. We used its training set, which contains 2394 tweets and 276 places.
     
    \item \textbf{MSM2013: \footnote{\url{https://www.researchgate.net/profile/Andrea-Varga-4/publication/256682215\_MSM2013\_Concept_Extraction\_Challenge\_dataset}}} It is the  gold dataset of Concept Extraction Challenge held at the Making Sense of Microposts Workshop in 2013 ( \#MSM2013). Entities of \textit{Person}, \textit{Organization}, \textit{Location}, and \textit{MISC} were annotated. We used its training set, which contains 2815 tweets and 619 places.
    %  \item \textbf{Datasets $q$ and $r$}: Many public spaces (e.g., hospital, park, and university) are tagged as \textit{Facility}, which however will be recognized as \textit{Location} by many location extractors. To mitigate this issue, we define that the detection of an entity which is \textit{Facility} in datasets $q$ and $r$ is a true positive. Moreover, datasets $q$ and $r$ normally ignore the location in hashtags, such as \textit{`Marysville'} in \textit{`\#MarysvilleShooting'}, which however can be recognized by some extractors. Thus, we define that, the detection of an entity that is not annotated in datasets $q$ and $r$ but in their mention part is not a false positive.

   \item  \textbf{WNUT2016 \footnote{\url{https://metatext.io/datasets/wnut-2016}}:} It is the gold data of the shared task on named entity recognition in Twitter. The task is part of the 2nd Workshop on Noisy User-generated Text (W-NUT 2016). Ten types of entities were annotated, such as \textit{Location}, \textit{Facility}, \textit{Person}, and \textit{Movie}.  We used its training set, which contains 3850 tweets and 602 places.

% \item \textbf{SemEva2019}: It is the gold dataset of the Task 12 (Toponym Resolution in Scientific Papers) of the 13th International Workshop on Semantic Evaluation (SemEva) \citep{weissenbacher2019semeval}. 150 full-text journal articles were downloaded from the subset of PubMed Central (PMC) and toponyms in the articles were manually annotated and linked to entries in GeoNames.

\item \textbf{LGL \footnote{\url{https://github.com/milangritta/Pragmatic-Guide-to-Geoparsing-Evaluation/blob/master/data/Corpora/lgl.xml}}:} Local-Global Lexicon (LGL) corpus was created by \citet{lieberman2010geotagging}. Toponyms were manually annotated and geocoded from 588 human-annotated news articles published by 78 local newspapers.

\item \textbf{GeoVirus \footnote{\url{https://github.com/milangritta/Pragmatic-Guide-to-Geoparsing-Evaluation/blob/master/data/Corpora/GeoVirus.xml}}:}
GeoVirus was created by \citet{gritta2018melbourne} for the evaluation of geoparsing approaches in news related to disease outbreaks and epidemics, such as Ebola, Bird Flu, and Swine Flu. Toponyms were manually annotated and geocoded. Only admit units were annotated in the dataset. Buildings, POIs, streets, and rivers were ignored.

\item \textbf{TR-News \footnote{\url{https://github.com/milangritta/Pragmatic-Guide-to-Geoparsing-Evaluation/blob/master/data/Corpora/TR-News.xml}}:} TR-News was created by \citet{kamalloo2018coherent}. Toponyms were manually annotated and geocoded from 118 news articles from various news sources.
 \end{itemize}

\begin{table}[]
\caption{Summary of 26 datasets. There are in total 39,736 places.}
\label{dataset}
\footnotesize
\begin{tabular}{|c|c|c|c|c|c|c|}
\hline
Name                                                         & Source & Type & \begin{tabular}[c]{@{}c@{}}Tweet \\ (Article) \\  Count\end{tabular} & \begin{tabular}[c]{@{}c@{}}Place \\ Count\end{tabular} & Resolved & Description                                                                                                                               \\ \hline
LaFlood2016 \citep{al2017location}          & tweet  & LE   & 1500                                                                 & 2295                                                   & No       & Louisiana flood in 2016                                                                                                                   \\ \hline
HouFlood2015 \citep{al2017location}         & tweet  & LE   & 1500                                                                 & 3060                                                   & No       & Houston flood in 2015                                                                                                                     \\ \hline
CheFlood2015 \citep{al2017location}         & tweet  & LE   & 1500                                                                 & 3671                                                   & No       & Chennai flood in 2015                                                                                                                     \\ \hline
Harvey2017 \citep{wang2020neurotpr}         & tweet  & LE   & 1000                                                                 & 2107                                                   & No       & 2017 Hurricane Harvey in  Texas and Louisiana                                                                                             \\ \hline
NzEq2013 \citep{middleton2018location}      & tweet  & LE   & 1994                                                                 & 1252                                                   & No       & New York hurricane  in 2012                                                                                                               \\ \hline
NyHurcn2012 \citep{middleton2018location}   & tweet  & LE   & 1997                                                                 & 764                                                    & No       & New Zealand  earthquake in 2013                                                                                                           \\ \hline
Martinez\_I  \citep{martinez2020knowledge}  & tweet  & LE   & 800                                                                  & 539                                                    & No       & Multiple emergency events across the world                                                                                                \\ \hline
Martinez\_II \citep{martinez2020knowledge}  & tweet  & LE   & 1371                                                                 & 642                                                    & No       & Multiple emergency events across the world                                                                                                \\ \hline
Martinez\_III \citep{nLORE}                 & tweet  & LE   & 8063                                                                 & 5122                                                   & No       & Multiple emergency events across the world                                                                                                \\ \hline
CrisisBench-1000  \citep{gazpne2}                                             & tweet  & LE   & 1000                                                                 & 1600                                                   & No       & 1000 tweets from  CrisisBench \citep{alam2020crisisbench}                                                                \\ \hline
HumAID-1000 \citep{gazpne2}                                                 & tweet  & LE   & 1000                                                                 & 1500                                                   & No       & 1000 tweets from HumAid \citep{humaid2020}                                                                               \\ \hline
COVID19-1000   \citep{gazpne2}                                                & tweet  & LE   & 1000                                                                 & 800                                                    & No       & 1000 tweets from   COVID-19  \citep{lamsal2021design}                                                                    \\ \hline
GeoCorpora \citep{wallgrun2018geocorpora}   & tweet  & LE   & 6634                                                                 & 3083                                                   & Yes      & Multiple  events across the world                                                                                                         \\ \hline
BTC-A \citep{derczynski2016broad}           & tweet  & NER  & 2000                                                                 & 229                                                    & No       & Section A of  Broad Twitter Corpus                                                                                                        \\ \hline
BTC-B \citep{derczynski2016broad}           & tweet  & NER  & 200                                                                  & 148                                                    & No       & Section B of Broad Twitter Corpus                                                                                                         \\ \hline
BTC-E \citep{derczynski2016broad}           & tweet  & NER  & 2000                                                                 & 572                                                    & No       & Section E of  Broad Twitter Corpus                                                                                                        \\ \hline
BTC-F \citep{derczynski2016broad}           & tweet  & NER  & 2113                                                                 & 1330                                                   & No       & Section  F of Broad Twitter Corpus                                                                                                        \\ \hline
BTC-G \citep{derczynski2016broad}           & tweet  & NER  & 1999                                                                 & 287                                                    & No       & Section G of  Broad Twitter Corpus                                                                                                        \\ \hline
BTC-H \citep{derczynski2016broad}           & tweet  & NER  & 1000                                                                 & 119                                                    & No       & Section H of   Broad Twitter Corpus                                                                                                       \\ \hline
NEEL2016 \citep{rizzo2015making}            & tweet  & NER  & 2135                                                                 & 602                                                    & Yes      & \begin{tabular}[c]{@{}c@{}}Dataset of Named Entity rEcognition and Linking\\  Challenge in 2016\end{tabular}                              \\ \hline
Ritte's dataset \citep{ritter2011named} & tweet  & NER  & 2394                                                                 & 276                                                    & No       & A general-purpose  NER dataset initially used in  \citep{ritter2011named}                                                                                                                          \\ \hline
MSM2013 \citep{cano2013making}              & tweet  & NER  & 2815                                                                 & 619                                                    & No       & \begin{tabular}[c]{@{}c@{}}Dataset of Concept Extraction Challenge at the \\ Making Sense of Microposts Workshop in 2013\end{tabular}     \\ \hline
WNUT2016 \citep{strauss2016results}         & tweet  & NER  & 3850                                                                 & 602                                                    & No       & \begin{tabular}[c]{@{}c@{}}Dataset of shared task on NER in Twitter at the \\  Workshop on Noisy User-generated Text in 2016\end{tabular} \\ \hline
LGL \citep{lieberman2010geotagging}        & news   & LE   & 588                                                                  & 5050                                                   & Yes      & Local-Global Lexicon corpus                                                                                                               \\ \hline
GeoVirus \citep{gritta2018melbourne}   & news   & LE   & 229                                                                  & 2167                                                   & Yes      & WikiNews related to global disease and epidemics                                                                                          \\ \hline
TR-News \citep{kamalloo2018coherent}    & news   & LE   & 118                                                                  & 1300                                                   & Yes      & Annotated news articles from various news sources                                                                                         \\ \hline
\end{tabular}

\end{table}

    We adopted the standard comparison metrics: precision, recall, and F1-score. In the case of overlapping or partial matches, we penalize a tool by adding 1/2 FP (False Positive) and 1/2 FN (False Negative) (e.g., if the tool marks \textit{`The Houston'} instead of \textit{`Houston'}), following \citet{al2017location}.

	\begin{equation}
		\label{formula: Min-Max Normalization}
Precision = \frac{TP}{TP+FP} 
\end{equation}

	\begin{equation}
Recall = \frac{TP}{TP+FN} 
\end{equation}

	\begin{equation}
F1 =  2 \cdot \frac{Precision \cdot Recall}{Precision + Recall}
\end{equation}

	  \subsection{Results of location reference recognition}
	  \label{sec:results}
	  
 We run the \toolss\thinspace tools on all the test datasets, and their precision, recall, and F1 score are reported in Figure \ref{result}. The three metrics are obtained by calculating the sum of FP, FN, and TP of all the datasets rather than by averaging the three metrics of the datasets due to the imbalanced place name count (from 119 to 5122) of different datasets. The raw result of the tools on each dataset can be seen from  \footnote{\url{https://docs.google.com/spreadsheets/d/16cyuyDhty04hQE1gBfP4zq23Lr3OWJ5EY\_bsVMoYxfQ/edit##gid=1536784545}}. %GazPNE2 achieves the best F1 score of 0.78. 
 The majority of tools can achieve very high precision with 21 of \toolss\thinspace achieving a precision over 0.7. Conversely, most of them obtain very low recall with only 2 (GazPNE2 and LORE) of \toolss\thinspace achieving a recall over 0.7. This means that most of the tools missed many location references. Therefore, we further investigate the reason in Section \ref{error} by analyzing the detection accuracy of the tools in different types of texts and different types of location references. We can also observe that the top five best performing tools, GazPNE2, Flair NER (Ont), nLORE, Flair NER, and Stanza, are all based on deep learning techniques and were proposed in the last four years, showing the superior performance and progress of deep learning on this task. Furthermore, two voting-based systems, SPENS and RS also achieve impressive results by simply combining the detection results of several classic tools, improving the performance of each single tool. The two voting systems are comparable with Stanza and Flair NER. This suggests a great potential of using voting mechanism in location reference recognition.%, considering the numerous available place name extractors. 
    
	  	   \begin{figure}[htbp!]
		\centering
		\includegraphics[width=0.98\textwidth]{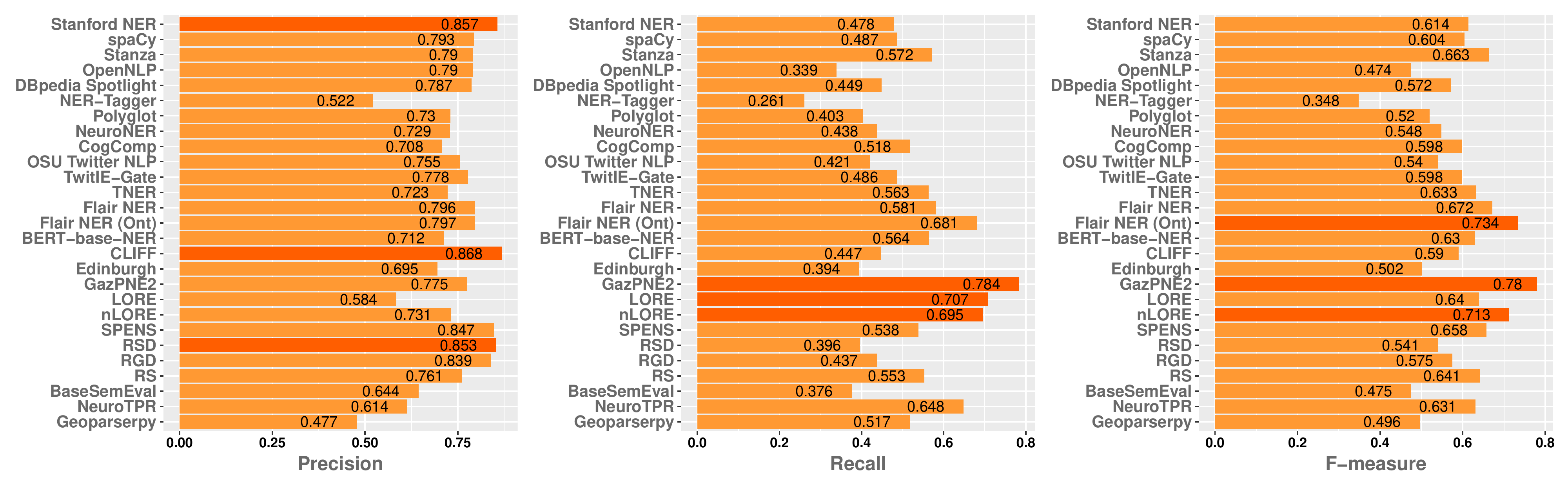}
		\caption{Precision, recall, and F1 score of the tested tools on all the datasets with  39,736 location references.}
		\label{result}
	\end{figure}

\subsection{Error analysis}
\label{error}
To gain an insight into what mistakes these tools made, we carried out an error analysis. Specifically, we investigated the performance of the tools on formal and informal texts. We also examined their performance on location references in different categories and forms.

\subsubsection{Text type}
Among the test datasets, 3 datasets are formal texts and 23 datasets are informal texts, containing 8517 and 31,219 places, respectively. The performance of the tools on the datasets of formal and informal texts is shown in Figure \ref{formal} and Figure \ref{informal}, respectively. We can observe that on formal texts, Flair NER (Ont), Flair NER, SPENS, Stanford NER, and Stanza perform the best, while on informal texts, GazPNE2 and nLORE perform the best. The main reason is that the former five tools are all trained on formal texts. Conversely, the latter two tools were proposed to deal with tweets and thus work better on twitter datasets. %Therefore, we suggest choosing different tools according to the type of texts.

	 \begin{figure}[htbp!]
		\centering
		\includegraphics[width=0.98\textwidth]{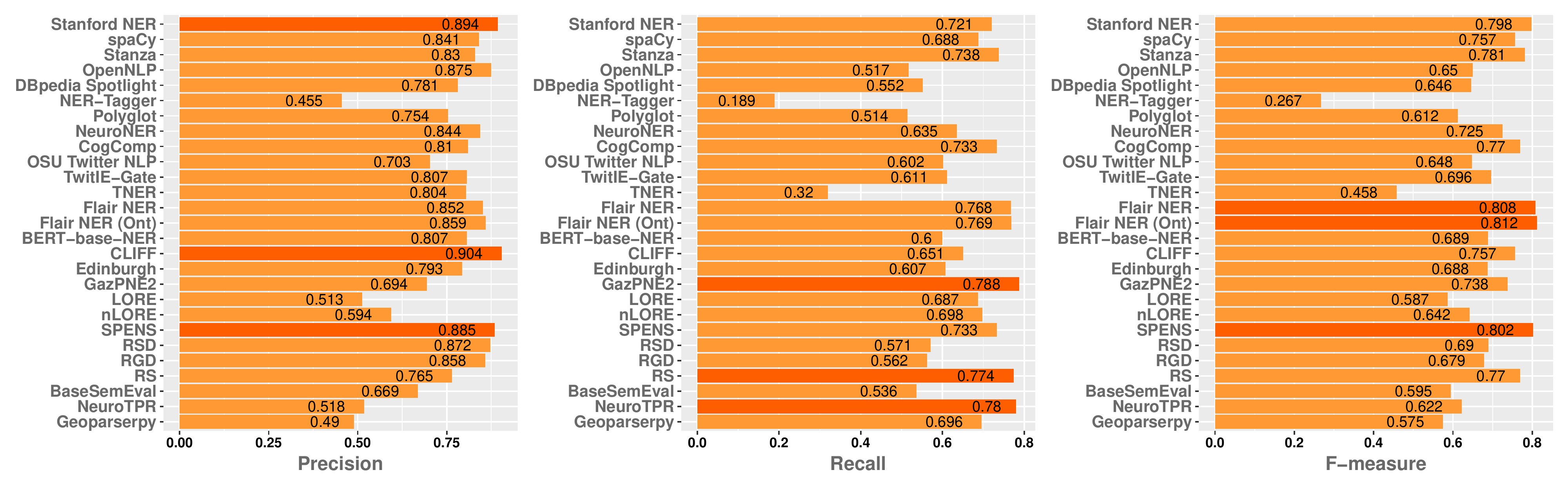}
		\caption{Precision, recall, and F1 score of the tested tools on the datasets of formal texts, containing 8517 places.}
		\label{formal}
	\end{figure}

	 \begin{figure}[htbp!]
		\centering
		\includegraphics[width=0.98\textwidth]{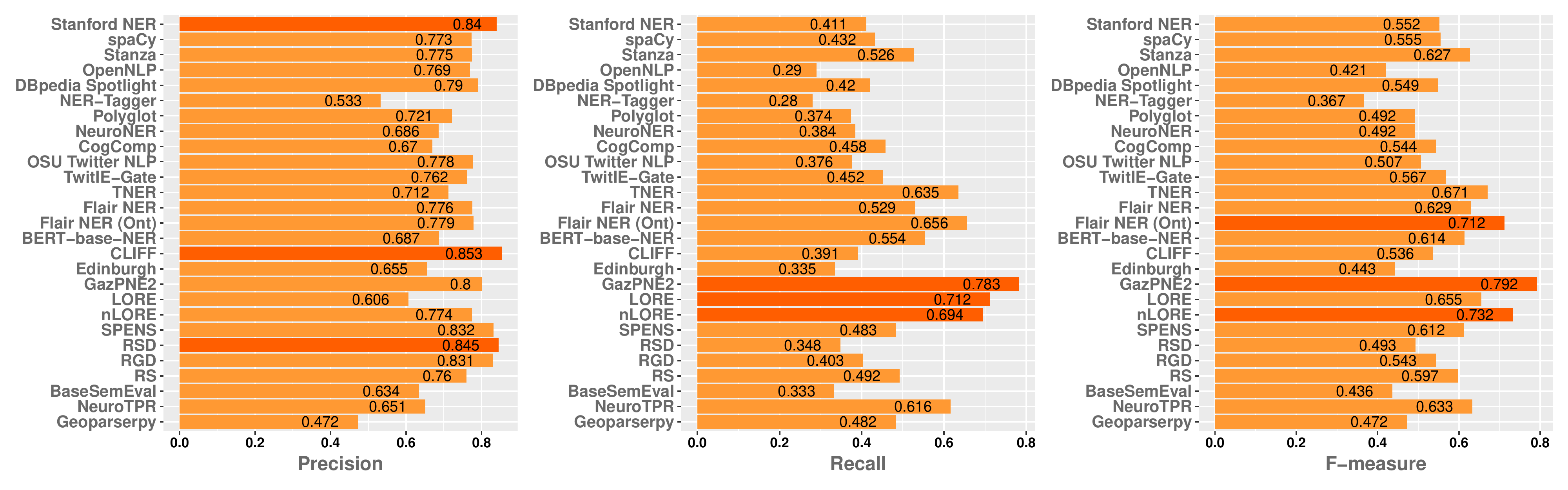}
		\caption{Precision, recall, and F1 score of the tested tools on the datasets of informal texts, containing 31,219 places.}
		\label{informal}
	\end{figure}

\subsubsection{Place category}
We divide the location references in the datasets into four categories: admin units (e.g., country, state, town, and suburb), traffic ways (e.g., street, road, highway, and bridge), natural features (e.g., river, creek, beach, and hill), and POIs (e.g., park, church, school, and library). We choose four datasets (i.e., Harvey2017, GeoCorpora, LGL, and TR-News) to conduct this experiment because the category of the places in the four datasets can be derived. The places in Harvey2017 were categorized into ten types \citep{hu2020people}. We treat the types of house number addresses, street names, highway, exits of highways, and intersections of roads as traffic ways, the type of natural features as natural features, the types of other human-made features and local organizations as POIs, and the types of admin units and multiple areas as admin units. In the other three datasets, place names were linked to the entries of GeoNames. We treat the places whose feature codes \footnote{\url{http://www.geonames.org/export/codes.html}} are \textit{A} (e.g., country, state, and region) and \textit{P} (e.g., city and village) in GeoNames as admin units, the places whose feature codes are \textit{R} (e.g., road and railroad) as traffic ways, the places whose feature codes are \textit{H} (e.g., stream and lake), \textit{T} (e.g., mountain, hill, and rock), \textit{U} (e.g., undersea and valley), \textit{V} (e.g., forest and grove) as natural features, and the places whose feature codes are \textit{L} (e.g., park and port) and \textit{S} (e.g., sport, building, and farm) as POIs. %Examples of \textit{A} are country, state, and region. Examples of \textit{P} are city and village. Examples of \textit{R} are road and railroad.  Examples of \textit{H} are stream and lake. Examples of \textit{T} are mountain, hill, and rock. Examples of \textit{U} are undersea and valley. Examples of \textit{V} are forest and grove. Examples of \textit{L} are park and area. Examples of \textit{S} are . %Table 1 lists the number of places of each category in the four datasets. 
There are in total 9790 admin units, 773 traffic ways, 263 natural features, and 754 POIs in the four datasets. %Table 1 lists the count of different place categories in the three datasets. 

We define the detection rate as the proportion of correctly detected places among the total places of a certain category. Only an exact match is regarded as a correct detection. Figure \ref{category} shows the detection rate of the tools on the four categories. We can observe that many tools show superior performance in recognizing coarse-grained places, with 13 of \toolss\thinspace recognizing over 60\% of the admin units. However, most of them are incapable of recognizing fine-grained places. Only 2 of \toolss\thinspace , 6 of \toolss\thinspace , and 4 of \toolss\thinspace can recognize over 60\% of the traffic ways, nature features, and POIs, respectively. The three categories refer to much more precise geographic scopes than admin units and are thus valuable in many key applications, such as emergency rescue and traffic event detection. %Therefore, recognizing fine-grained places should be paid more attention. 
It is worth mentioning that GazPNE2 can recognize over 70\% of places in all four categories.

	\begin{figure}[htbp!]
		\centering
		\includegraphics[width=0.65\textwidth]{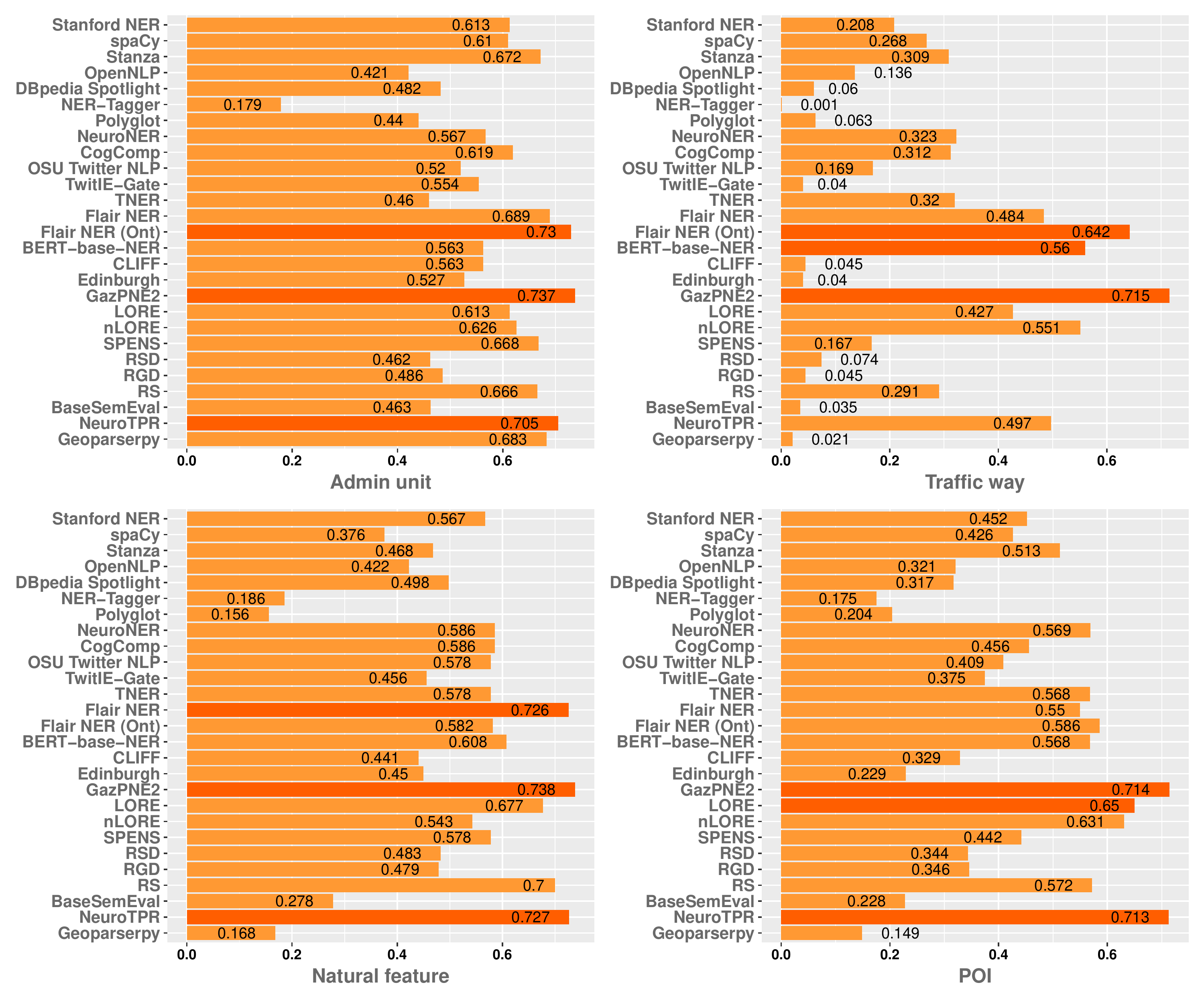}
		\caption{Detection rate of tools on the four categories with 9790 admin units, 773 traffic ways, 263 natural features, and 754 POIs.}
		\label{category}
	\end{figure}

%However, place names annotated in the datasets were not categorized. Therefore, we try to determine the category of the annotated place names by using a semi-automatic way. Specifically, we input distribution of length of place names in the datasets are shown in Figure 2.the place names in OpenStreetMap to find their matches. The categories of the place name are determined by the features of the matches on OSM. When conflicts appear or no matches are found, they will be manually checked by us. At last, we manually checked 6000 place names. 

% \subsubsection{Length of place names}
% The distribution of length of place names in the total datasets are shown in Figure 2. Figure  \ref{length} shows the detection rate of the tools on the place names of different length. 
% 	  	   \begin{figure}[htbp!]
% 		\centering
% 		\includegraphics[width=0.8\textwidth]{fig//result.png}
% 		\caption{Detection rate of tools on the place names of different length.}
% 		\label{length}
% 	\end{figure}
% 	We can see, most of the tools work well on one-word place names, but fail in detecting multi-word place names.

\subsubsection{Form of location references}
We consider three forms of location references: the place names with numbers (e.g., \textit{`500 Neches Ave'} and \textit{`Highway 25'} ), abbreviation of place names (e.g., \textit{`us'} and \textit{`tx'} ), and place names in hashtags (e.g., \textit{`\#\textbf{Houston}Flood'} and \textit{`\#\textbf{Chennai}'} ). Place names with numbers mostly refer to fine-grained locations, such as highways, roads, and home addresses.  % since many place name extractors fail to detect the three kinds of place names. 
We define that the abbreviation of place names consists of only one single word and its char length does not surpass 3. %The category of hashtag can be determined by the location of $\#$ in texts. 
%Notably, the type of abbreviation and hashtags might overlap. 
1621, 3697, and
6560 place names are in the three forms (number, abbreviation, and hashtag), respectively. Figure  \ref{form} shows the detection rate of the tools on the place names of the three forms.
	  	   \begin{figure}[htbp!]
		\centering
		\includegraphics[width=0.98\textwidth]{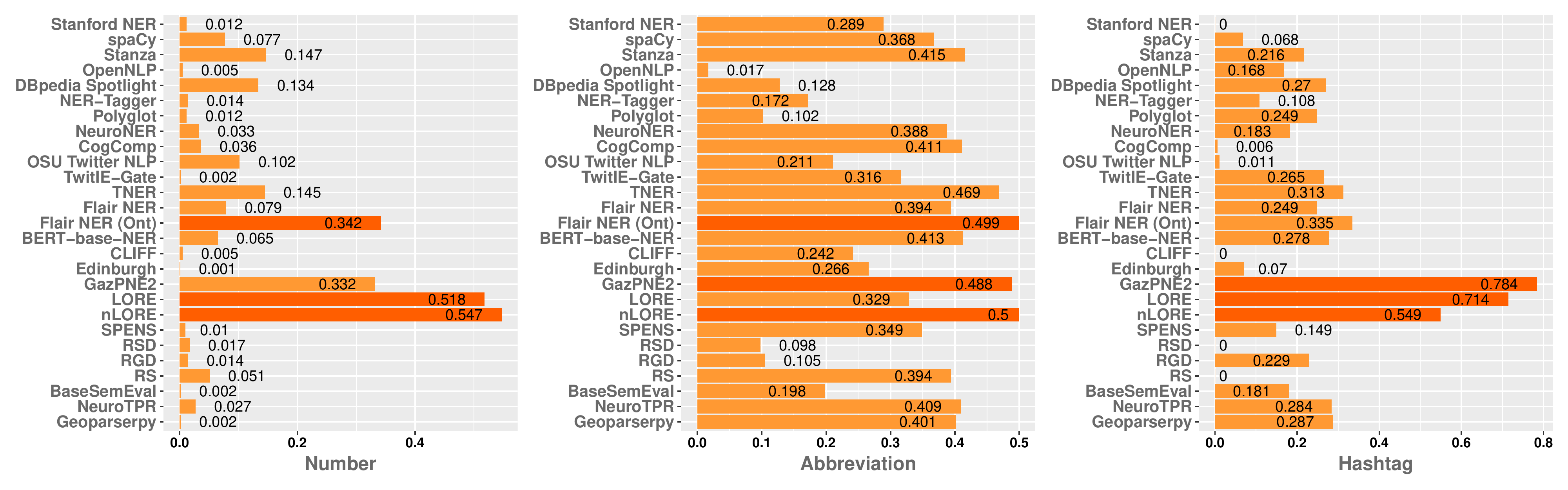}
		\caption{Detection rate of tools on location references in different forms with 1621 place names with numbers, 3697 abbreviations, and
6560 place names in hashtags.}
		\label{form}
	\end{figure}
%We can see, most of the tools work well on one-word place names, but fail in detecting multi-word place names.

We can observe that it is a challenge to recognize place names with numbers, with only 4 of \toolss\thinspace achieving a detection rate of over 0.3. Conversely, recognizing abbreviations is easier, with over half (16 of \toolss\thinspace) of the tools recognizing over 30\% of the abbreviations. However, we can also observe that no tool can achieve a detection rate over 0.6 on the two forms. It is also a challenge to recognize place names in hashtags, with only 5 of \toolss\thinspace achieving a detection rate of over 0.3. However, it is encouraging that GazPNE2 and LORE can recognize over 70\% of the places in hashtags. %Much more effort should be put on detecting place names with numbers and abbreviations of place names.

\subsection{Computational efficiency}
In this section, we further investigate the computational efficiency (i.e., speed) of different approaches. In many applications, the texts that need to be geoparsed are of huge volumes, such as some major historical books and reports (e.g., the Old Bailey Online) that each comprise many millions or even billions of words \citep{gregory2015geoparsing} and over millions of tweets related to a crisis event \citep{qazi2020geocov19}. This requires a rapid geoparsing procedure and the speed is thus a critical indicator. %That is, apart from accuracy, speed is another important indicator that decides if an approach should be adopted in a certain application. 

We run each approach on the total datasets and record the consumed time of each approach. We do not count the consumed time during the training phase if an approach needs to be trained since it can be conducted offline and is a one-time process. Note that, we do not include Edinburgh Geoparser and DBpedia Spotlight in the comparison since they are online service and it is impossible to count their amount of time of processing  done on the server. While most of approaches run on a MacBook Pro laptop with an Intel Core i7 (2.2 GHz 6-Core) and a RAM of 16 GB, three approaches (i.e., OSU TwitterNLP, LORE, and nLORE) run on a Lenovo laptop with an Intel Core i5 (2.5 GHZ 4-Core) and a RAM of 3.8 GB since they require a Linux or Windows environment. Figure \ref{computation} illustrates the speed of the approaches.

	  	   \begin{figure}[htbp!]
		\centering
		\includegraphics[width=0.7\textwidth]{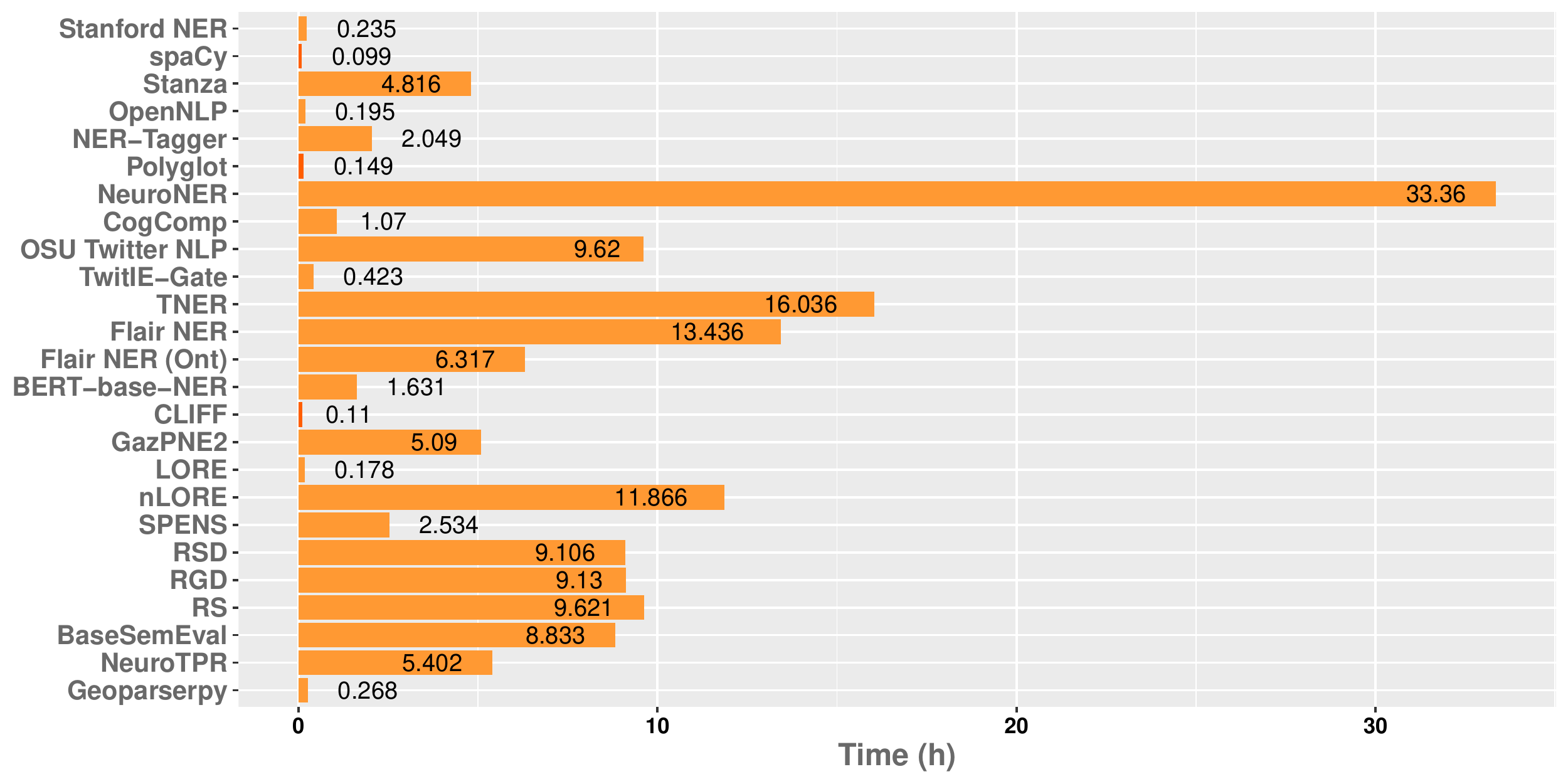}
		\caption{Time consumption of the approaches running on the total test datasets.}
		\label{computation}
	\end{figure}

We can observe that the speed of different approaches varies drastically. It takes 6 minutes to 33 hours for these approaches to process the total datasets that contain 1,092,093 words. It is unexpected that OSU Twitter NLP takes nearly 9.6 hours. Therefore, RSD, RGD, and RS that use the OSU Twitter NLP take also nearly 10 hours. The other approaches that take over 5 hours are all deep learning-based. spaCy, Cliff, LORE, Ployglot, and OpenNLP are 20 times quicker than these approaches. However, the deep learning-based approaches achieve a much higher F1 score than the other approaches. Therefore, there exists a trade-off between correctness and computational efficiency.

\section{conclusions and outlook}
In this paper, we first summarized seven typical applications of geoparsing, and then presented a  survey of existing approaches for location reference recognition. We grouped these existing approaches into four categories: rule-based, gazetteer matching-based, statistical learning-based, and hybrid approaches. In addition, we carried out a  comparative study to systematically compare \toolss\thinspace existing approaches on 26 datasets. We evaluated these approaches from multiple perspectives including their overall location recognition accuracy, their performance on formal (e.g., news) and informal (e.g., tweets) texts, and their capabilities to detect location references in different categories and forms. Finally, we also compared the computational efficiency of these existing approaches.

%using 23 twitter datasets and 3 datasets of formal texts,  from multiple dimensions, including their performance on  texts and their detection rate on places of different attributes (category and form), and their computational efficiency (i.e., speed). 

From the results, we can conclude that: (1) deep learning is so far the most promising technique in location reference recognition; %, and should be paid more attention to; 
(2) fusing existing approaches or tools in a voting mechanism can overcome each others' shortcomings and is more robust than each single approach; %The idea can be extended to other domains %, such as to vote the results of multiple geocoders
(3) the performance of different approaches varies on the type of texts and the attributes of location references and their computational efficiency also vary drastically. Users should choose the proper one according to their specific application demands. For example, for traffic management and disaster management that rely on informal texts and require fine-grained locations, GazPNE2 is a good option; for crime management and tourism management that rely on formal texts and require both coarse-grained and fine-grained locations, Flair NER (Ont) is a good option; for spatial humanities, disease surveillance, and GIR that need process a large number of formal texts and require just coarse-grained locations, Standford NER, CogComp, and spaCy are good options. %(4) , with the consumed time ranging from 1 hour to 1 day on the total dataset that comprises around one million words. Speed is another important indicator that should be taken into account when selecting an approach for a specific application. 

Several research directions can be further explored in the future. 
\begin{itemize}
    \item \textbf{Location reference recognition}: There is still space to improve the performance of the approaches for informal texts (e.g., tweets) since most of the existing approaches did not perform well on informal texts. Inspired by the success of two voting systems RS and SPENS, one of the promising ways might be further selecting several approaches from the \toolss\thinspace approaches and combining them in a voting system to achieve more satisfying results. We can also configure the combined approaches  of the voting system to satisfy the requirements of different applications, such as a high recall on POIs of tweets. %The voting mechanism can be also applied to extract place names from texts in other languages.

\item \textbf{Location reference geocoding}: While there exist many approaches for geocoding location references, they focused mainly on formal texts. Some studies proposed geoparsing tweets, but usually limited the application area to a known geographic region (e.g., a city where a flood occurs) \citep{ahmed2019real,al2017location}. In those cases, simply searching a local gazetteer is often sufficient for geocoding. Only a few studies \citep{karimzadeh2019geotxt,nizzoli2020geo} geoparsed tweets at a global scale. This task has two main challenges: geo/geo ambiguities caused by limited contexts in short texts of tweets and unseen place names caused by place name variants and the informal features of tweets that might contain abbreviations, slang, and misspellings. Three main ways might be explored to overcome the challenges: (1) to leverage deep learning \citep{fize2021deep,cardoso2021novel,kulkarni2021multi} and clustering techniques \citep{de2018taggs} that can group tweets of the same topic to expand the context of tweets; (2) to combine multiple SOTA geocoders in a voting mechanism; (3) to generate abundant and high-quality training examples by using geotagged tweets. The geotag of a tweet can drastically remove the geo/geo ambiguities of the place names in the tweet text, enabling the generation of high-quality training examples. %To overcome the first challenge, clustering methods \citep{de2018taggs} that can group similar tweets should be investigated. To overcome the second challenge, spatial embedding techniques \citep{di2021sherloc} should be given more attention.

\item \textbf{Datasets for geoparsing research}: Existing datasets for geoparsing research are often in the form of formal text such as news, and only a few twitter datasets (e.g., GeoCorpora and NEEL2016) are available for geoparsing research. Although there exist many other twitter datasets for general NER research, these datasets typically do not contain geographic coordinates for the labeled entities and therefore cannot support research involving the entire workflow of geoparsing. More datasets in informal texts with labeled location references and their geographic coordinates are needed. Furthermore, most of the location references in the existing datasets are admin units, such as countries and cities. Finer-grained location references, such as traffic ways and POIs, are much rare. However, they are important in many applications, such as determining the precise locations where rescue is needed during disasters. A large twitter dataset that contains many fine-grained locations across the world would be very helpful for advancing methods in recognizing and geocoding fine-grained location references in texts.

\end{itemize}

\bibliographystyle{ACM-Reference-Format}
\bibliography{sample-base}

\appendix

\end{document}